\documentclass[orivec]{llncs}

\usepackage{times}
\usepackage{graphicx}
\usepackage{algorithmicx}
\usepackage{algorithm}
\usepackage{algpseudocode}
\usepackage{longtable}
\usepackage{amsmath}
\usepackage{amssymb}

\usepackage{url}
\usepackage[colorinlistoftodos]{todonotes}
\usepackage{paralist}
\usepackage{xspace}
\usepackage{subcaption}
\usepackage{lipsum}
\usepackage{cite}
\usepackage{bm}
\usepackage{paralist}

 \usepackage{multicol} 
 \usepackage{multirow} 

\usepackage{lipsum}

\usepackage[shortlabels]{enumitem}
\setlist[enumerate]{nosep}

\usepackage{xspace}

\newcommand{\ie}{\textit{i}.\textit{e}.,\xspace}
\newcommand{\eg}{\textit{e}.\textit{g}.,\xspace}

\newcommand{\ontoOne}{\ensuremath{\O_1}}
\newcommand{\ontoTwo}{\ensuremath{\O_2}}

\newcommand{\FMA}{\ensuremath{\O_\text{FMA}}}
\newcommand{\NCI}{\ensuremath{\O_\text{NCI}}}

\newcommand{\ontoOneP}[1]{\ensuremath{\ontoOne}^{#1}}
\newcommand{\ontoTwoP}[1]{\ensuremath{\ontoTwo}^{#1}}

\renewcommand{\O}{\ensuremath{\mathcal{O}}\xspace}
\newcommand{\M}{\ensuremath{\mathcal{M}}}
\newcommand{\T}{\ensuremath{\mathcal{T}}}

\newcommand{\D}{\ensuremath{\mathcal{D}}}
\newcommand{\mapping}[4]{\langle #1,\allowbreak #2,\allowbreak #3,\allowbreak #4
\rangle}
\newcommand{\mappingTwo}[2]{\langle #1,\allowbreak #2 \rangle}
\newcommand{\myset}[1]{ \{#1\} }
\newcommand{\MS}{\M^{S}\xspace}
\newcommand{\MRA}{\M^{RA}\xspace}
\newcommand{\MRAFN}{\MRA_{\text{fma-nci}}\xspace}

\newcommand{\MLex}{\M^{\lex}\xspace}
\newcommand{\MT}{\ensuremath{\M\T}\xspace}
\newcommand{\MTLex}{\MT^{\lex}\xspace}
\newcommand{\PMT}{\ensuremath{\D_{\MT}}\xspace}

\newcommand{\MTMO}{\MT^{\M}_{\O_1\text{-}\O_2}\xspace}
\newcommand{\MTRAFN}{\MT^{RA}_{\text{fma-nci}}\xspace}

\newcommand{\cem}[1]{\mathsf{#1}}
\newcommand{\nsp}{\negthickspace}

\newtheorem{hypothesis}{Hypothesis}

\newcommand{\lexN}{\textsf{LexI}}
\newcommand{\lex}{\textsf{LexI}\xspace}
\newcommand{\starspace}{\textsf{StarSpace}\xspace}

\newcommand{\logicalalgo}[1]{\ifmmode \text{ \textbf{#1} } \else \textbf{#1} \fi}

\newcommand{\funcalgo}[1]{\ensuremath{\mathsf{#1}}}

\begin{document}

\title{Breaking-down the Ontology Alignment Task\\ with a Lexical Index and
Neural Embeddings}

\author{Ernesto Jim\'enez-Ruiz\inst{1,2},
Asan Agibetov\inst{3},
Matthias Samwald\inst{3},
Valerie Cross\inst{4}
}

\institute{    
    The Alan Turing Institute, London, United Kingdom
    \and 
    Department of Informatics,
    University of Oslo, Norway
    \and 
    Section for Artificial Intelligence and Decision Support, Medical University of Vienna, Vienna, Austria
    \and
   	Miami University, Oxford, OH 45056, United States
}

\maketitle

\begin{abstract} 
Large ontologies still pose serious challenges to state-of-the-art ontology
alignment systems. In the paper we present an approach that combines a lexical
index, a neural embedding model and locality modules to effectively 
divide an input ontology matching task into smaller and more tractable
matching (sub)tasks. We have conducted a comprehensive evaluation using the datasets of the Ontology Alignment Evaluation Initiative. The results are encouraging and suggest that the proposed methods are adequate in practice and can be
integrated within the workflow of state-of-the-art systems.
\end{abstract}


\setcounter{footnote}{0}

\section{Introduction}

The problem of (semi-)automatically computing an alignment between
independently developed ontologies has been extensively studied in the last
years \cite{DBLP:journals/tkde/ShvaikoE13, DBLP:books/daglib/0032976}. As a
result, a number of sophisticated ontology alignment systems currently
exist.\footnote{Ontology matching surveys and
approaches: \url{http://ontologymatching.org/}} The Ontology
Alignment Evaluation Initiative\footnote{OAEI evaluation campaigns:
\url{http://oaei.ontologymatching.org/}} (OAEI)
\cite{DBLP:conf/semweb/AchichiCDEFFFFH17} has played a key role in the benchmarking of these systems by
facilitating \begin{inparaenum}[\it (i)]
\item their comparison on the same basis, and 
\item the reproducibility of the evaluation and results.
\end{inparaenum}
The OAEI includes different tracks organised by different research groups.
Each track contains one or more matching tasks involving small-size (\eg
conference), medium-size (\eg anatomy), large (\eg phenotype) or
very large (\eg largebio) ontologies.

Large ontologies still pose serious challenges to ontology alignment systems.
For example, only 6 out of 21
participating systems in the OAEI 2017 campaign were able to complete the
largest tasks in the \emph{largebio track}
\cite{DBLP:conf/semweb/AchichiCDEFFFFH17}.
OAEI systems are typically able to cope with small and medium size
ontologies, but fail to complete large tasks in a given time frame and/or with
the available resources (\eg memory). 
Prominent examples across the OAEI campaigns are:
\begin{inparaenum}[\it (i)]
\item YAM++ version 2011 (best results in \emph{conference} track, but failed to
complete the \emph{anatomy} task);
\item CODI version 2011.5 (best results in \emph{anatomy} but could not cope with the
\emph{largebio} track); 
\item Mamba version 2015 (top system in the \emph{conference} track but could not complete
the \emph{anatomy} track); 
\item FCA-Map version 2016 (completed both \emph{anatomy} and
\emph{phenotype} tasks but did not complete the largest \emph{largebio} tasks);
and
\item POMap version 2017 (one of the top mappings in \emph{anatomy} but could not finish the
largest \emph{largebio} tasks).
\end{inparaenum}

Shvaiko and Euzenat \cite{DBLP:journals/tkde/ShvaikoE13} list some potential
solutions to address the challenges that large ontologies pose to ontology alignment systems, namely:
parallelization, distribution, approximation, partitioning and optimization.
%
%
%
%
%
In this paper we propose a novel method to effectively divide the matching
task into several (independent) smaller (sub)tasks. This method relies on
an efficient lexical index (as in LogMap \cite{logmap2012}), a
neural embedding model~\cite{wu_2017} and locality
modules~\cite{DBLP:journals/jair/GrauHKS08}.
Unlike other 
state-of-the-art approaches, 
our method 
provides guarantees about the preservation of the coverage of 
the relevant ontology alignments as defined in Section \ref{sect:quality}.

The remainder of the paper is organised as follows. Section
\ref{sec:preliminaries} introduces the main concepts that will be used in the
paper. Section \ref{sec:methods} presents the methods and strategies to divide the ontology matching task into a set of smaller subtasks.
The conducted evaluation is provided in Section
\ref{sec:eval}.
 Section \ref{sec:related} summarizes the related literature.
Finally, Section \ref{sec:disc} concludes the paper and suggests some lines of
immediate future research.


\section{Preliminaries}
\label{sec:preliminaries}

In this section we introduce 
the background concepts that will be
used throughout the paper. 

\subsection{Basic definitions}

A \emph{mapping} (also called \textit{match} or \textit{correspondence})
between entities\footnote{We refer to (OWL 2) classes, data and object properties and named individuals as entities.} of two ontologies\footnote{We assume ontologies are expressed in OWL 2 \cite{OWL2}.} $\O_1$ (\ie source) and $\O_2$ (\ie target) is typically represented as a 
4-tuple $\mapping{e_1}{e_2}{r}{c}$ where $e_1$ and $e_2$ are entities of $\O_1$ and $\O_2$, 
respectively; $r \in  \myset{\sqsubseteq, \sqsupseteq, \equiv}$ is a semantic relation; and $c$ is a confidence value, 
usually, a real number within the interval $\left(0,1\right]$. In our approach we simply consider mappings as a pair $\mappingTwo{e_1}{e_2}$
%
An ontology \emph{alignment} is a
set of mappings $\M$ between two ontologies $\O_1$ and $\O_2$.

An ontology \emph{matching task} $\MT$ is composed of a pair of ontologies
$\O_1$ and $\O_2$ and possibly an associated \emph{reference alignment} $\MRA$.
The objective of a matching task is to discover an overlapping of $\O_1$ and
$\O_2$ in the form of an alignment $\M$. The \emph{size or search space} of a
matching task is typically bound to the size of the 
Cartesian product between the entities of the input ontologies:
$\lvert Sig(\O_1) \rvert \times \lvert Sig(\O_2)
\rvert$ being $Sig(\O)$ the signature (\ie entities) of $\O$. 

An ontology \emph{matching system} is a program
that, given as input the ontologies $\O_1$ and $\O_2$ of a matching task,
generates an ontology alignment $\MS$.


The standard evaluation measures for an alignment $\MS$ are
\emph{precision} (P), \emph{recall} (R) and \emph{f-measure} (F) computed against a reference alignment $\MRA$ as follows:
\begin{equation}\label{eq:measures}
    P = \frac{\lvert\MS \cap \MRA\rvert}{\lvert\MS\rvert},~
    R = \frac{\lvert\MS \cap \MRA\rvert}{\lvert\MRA\rvert},~
    F = 2 \cdot \frac{P \cdot R}{P + R} 
\end{equation}

\subsection{Matching subtasks and quality measures: size ratio and coverage}
\label{sect:quality}

We denote \emph{division} of an ontology matching task $\MT$, composed by
the ontologies $\O_1$ and $\O_2$, as the process of finding
matching subtasks $\MT_{i}=\langle \ontoOne^{i}, \ontoTwo^{i} \rangle$ (with
$i$=$1$,\ldots,$n$), where $\ontoOne^{i} \subset \ontoOne$ and $\ontoTwo^{i} \subset
\ontoTwo$. 
%
The size of the matching subtasks aims at being smaller than the
original task in terms of search space. 
Let $\PMT^{n}=\{\MT_1,\ldots,\MT_n\}$ be the result of dividing a matching task $\MT$, the
\emph{size ratios} of the
matching subtasks $\MT_i$ and $\PMT^{n}$
are
computed as follows:

%
\begin{equation}\label{eq:sizeeratioTask}
    \funcalgo{SizeRatio}(\MT_i, \MT) = \frac{\lvert Sig(\ontoOne^{i}) \rvert
    \times \lvert Sig(\ontoTwo^{i}) \rvert} {\lvert Sig(\O_1) \rvert \times \lvert
    Sig(\O_2)\rvert}
\end{equation}

\begin{equation}\label{eq:sizeeratio}
    \funcalgo{SizeRatio}(\PMT^{n}, \MT) =
    \sum_{i=1}^{n}\funcalgo{SizeRatio}(\MT_i, \MT)
\end{equation}

The ratio $\funcalgo{SizeRatio}(\MT_i, \MT)$ is expected to be less than
$1.0$ while the aggregation $\sum_{i=1}^{n}\funcalgo{SizeRatio}(\MT_i, \MT)$,
being $n$ the number of matching subtasks, can be greater than
$1.0$ (as matching subtasks may overlap), that is, the aggregated size of the
matching subtasks may be larger than the original task size in
terms of (aggregated) search space.

The \emph{coverage} of the matching subtask aims at providing guarantees about the preservation of the (potential) outcomes of the original matching task. That is, it indicates if the relevant ontology alignments in the original matching task can still be computed with the matching subtasks. The coverage is
calculated with respect to a relevant alignment $\M$, possibly the
reference alignment $\MRA$ of the matching task if it exists. The formal notion
of coverage is given in Definitions \ref{def:cov1} and \ref{def:cov2}. 

\begin{definition}[Coverage of a matching task]
\label{def:cov1}
Let $\MT=\langle \ontoOne, \ontoTwo \rangle$ be a matching task and $\M$ an
alignment. We say that a mapping $m=\mappingTwo{e_1}{e_2} \in \M$ is covered by
the matching task if $e_1 \in Sig(\O_1)$ and $e_2 \in Sig(\O_2)$. The
coverage of $\MT$ w.r.t. $\M$ (denoted as $\funcalgo{Coverage}(\MT, \M)$) 
represents the set of mappings 
$\M' \subseteq \M$ 
covered~by~$\MT$.
\end{definition}

\begin{definition}[Coverage of the matching task division]
\label{def:cov2}
Let $\PMT^{n}=\{\MT_1,\ldots,\MT_n\}$ be the result of dividing a matching
task $\MT$ and $\M$ an alignment. We say that a mapping $m\in
\M$ is covered by $\PMT$ if $m$ is at least covered by one of the matching
subtask $\MT_i$ (with $i$=$1$,\ldots,$n$) as in Definition \ref{def:cov1}.
The coverage of $\PMT$ w.r.t. $\M$ (denoted as $\funcalgo{Coverage}(\PMT, \M)$) 
represents the set of mappings $\M' \subseteq \M$ covered by $\PMT$.
The coverage will typically be given as a ratio with respect to the (covered)
alignment:

\begin{equation}\label{eq:covratio}
    \funcalgo{CoverageRatio}(\PMT^{n}, \M) =
    \frac{\lvert\funcalgo{Coverage}(\PMT, \M)\rvert} {\lvert \M \rvert}
\end{equation}
\end{definition}


\subsection{Locality-based modules in ontology alignment}

Logic-based module extraction techniques compute ontology fragments that capture
the meaning of an input signature 
with respect to a given
ontology. In this paper we rely on bottom-locality modules
\cite{DBLP:journals/jair/GrauHKS08}, which will be referred to as locality-modules or simply as modules.
Locality modules play an important role in ontology alignment tasks. For
example they provide the scope or context (\ie sets of \emph{semantically related} entities \cite{DBLP:journals/jair/GrauHKS08}) 
for the entities in a given mapping or set of mappings as formally
presented in Definition~\ref{def:context1}.

\begin{definition}[Context of a mapping and an alignment]
\label{def:context1}
Let $m=\mappingTwo{e_1}{e_2}$  be a mapping
between two ontologies $\O_1$ and $\O_2$. We define the context of $m$ (denoted
as $\funcalgo{Context}(m, \O_1, \O_2)$) as a pair of modules $\ontoOne'
\subseteq \O_1$ and $\ontoTwo' \subseteq \O_2$, where $\ontoOne'$ and $\ontoTwo'$ include
the semantically related entities to $e_1$ and $e_2$, respectively \cite{DBLP:journals/jair/GrauHKS08}.
Similarly, the \emph{context} for an alignment $\M$ between two ontologies
$\O_1$ and $\O_2$ is denoted as $\funcalgo{Context}(\M, \O_1, \O_2)=\langle
\ontoOne', \ontoTwo' \rangle$, where  $\ontoOne'$ and $\ontoTwo'$ are modules
including the semantically related entities for the entities $e_1 \in
Sig(\O_1)$ and $e_2 \in Sig(\O_2)$ in each mapping $m=\mappingTwo{e_1}{e_2} \in \M$.
\end{definition}

\subsection{Context as matching task}

The context of an alignment between two ontologies represents the overlapping of these ontologies with respect to the aforesaid alignment. Intuitively, 
the ontologies in the context of an alignment will cover all the mappings in that alignment. 
Definition~\ref{def:overlappingMT} formally presents the context of an alignment as the \emph{overlapping} matching task to discover that alignment.


\begin{definition}[Overlapping matching task]
\label{def:overlappingMT}
Let $\M$ be an alignment between $\O_1$ and $\O_2$, and
$\funcalgo{Context}(\M, \O_1, \O_2)=\langle \ontoOne', \ontoTwo' \rangle$ the
context of $\M$. We define $\MTMO=\langle \ontoOne', \ontoTwo' \rangle$ as the overlapping matching task for $\M$. A matching task $\MT=\langle \ontoOne, \ontoTwo \rangle$ can be reduced to the task $\MTMO=\langle \ontoOne', \ontoTwo' \rangle$ without information loss in terms of finding $\M$.
\end{definition}

A matching system should aim at computing $\M$ with both the reduced task $\MTMO$ and the original matching task $\MT$.
For example, in the OAEI \emph{largebio} track~\cite{DBLP:conf/semweb/AchichiCDEFFFFH17} instead of the original matching task (\eg whole FMA and NCI ontologies), they are given the context of the reference alignment (\eg  $\MTRAFN = \funcalgo{Context}(\MRAFN,\
\FMA, \NCI)=\langle \FMA', \NCI' \rangle$) as a (reduced) overlapping matching task.






\section{Methods}
\label{sec:methods}

The approach presented in this paper relies on an `inverted'
lexical index (we will refer to this index as \lex), commonly  used  in  information  retrieval applications, and also
used in ontology alignment systems like LogMap~\cite{logmap2012} or
ServOMap~\cite{servomap14}.

\begin{table*}[t!]
\caption{Inverted lexical index \lex (left) and entity index (right). For readability,
stemming techniques have not been applied and index values have been split into elements of $\ontoOne$ and $\ontoTwo$. `-'
indicates that the ontology does not contain entities for that  
entry.}\label{table:ext_if}%
\vspace{-0.1cm}
\centering
{

\begin{tabular}{ll}
\begin{footnotesize}
\begin{tabular}[t]{|l||l|l|}
\hline 

\multirow{2}{*}{\textbf{Index key}} & \multicolumn{2}{c|}{\textbf{Index value}}
\\\cline{2-3} 
& \textbf{Entities $\ontoOne$} & \textbf{Entities $\ontoTwo$} \\\hline

$\{$ acinus $\}$ & 7661,8171 & 118081 \\\hline

$\{$ mesothelial, pleural $\}$ & 19987 & 117237 \\\hline
 \multicolumn{3}{c}{\tiny{~}\vspace{-0.21cm}} \\\hline

$\{$ hamate, lunate $\}$ & 55518 & - \\\hline

$\{$ feed, breast $\}$ & - & 113578,111023 \\\hline



\end{tabular}
\end{footnotesize}
&
\begin{scriptsize}
\begin{tabular}[t]{|l||l|}
\hline 
\textbf{ID} & \textbf{URI} \\\hline
7661 & \ontoOne:Serous\_acinus \\
8171 & \ontoOne:Hepatic\_acinus \\
19987 & \ontoOne:Mesothelial\_cell\_of\_pleura\\
55518 & \ontoOne:Lunate\_facet\_of\_hamate\\\hline
118081 & \ontoTwo:Liver\_acinus \\
117237 & \ontoTwo:Pleural\_Mesothelial\_Cell\\
113578 & \ontoTwo:Breast\_Feeding\\
111023 & \ontoTwo:Inability\_To\_Breast\_Feed\\\hline

\end{tabular}
\end{scriptsize}
\end{tabular}

}
\end{table*}

\subsection{The lexical index \lex}

\lex encodes the labels of all entities of the input ontologies $\ontoOne$ and $\ontoTwo$, including their lexical variations (\eg preferred labels, synonyms), in the form of pairs \emph{key-value}
where the key is a set of words and the value is a set of entity
identifiers\footnote{The indexation module associates unique numerical
identifiers to entity URIs.} such that the set of words of the key
appears in (one of) the entity labels. Table \ref{table:ext_if} shows a few example entries of \lex for two input ontologies.

\lex is created as follows.
\begin{inparaenum}[\it (i)]
  \item Each label associated to an ontology entity is split into a set of
  words; for example, the label ``Lunate facet of hamate'' is split into the set \{``lunate'', ``facet'', ``of'', ``hamate''\}.
  \item Stop-words are removed, for example,``of'' is removed from the set of words (\ie \{``lunate'', ``facet'', ``hamate''\}).  
  \item Stemming techniques are applied to each word (\ie \{``lunat'', ``facet'', ``hamat''\}). 
  \item Combinations of (sub)set of words serve as keys in \lex; for example,
  \{``lunat'', ``facet''\}, \{``hamat'', ``lunat''\} and so on.\footnote{In
  order to avoid a combinatorial blow-up the number of computed subsets of words
  is limited.}
  \item Entities leading to the same (sub)set of words are associated to the
  same key in \lex, for example, the entity $\cem{\ontoOne\nsp:\nsp Lunate\_facet\_of\_hamate}$ with numerical identifier 55518 is associated to the \lex key \{``hamat'', ``lunat''\} (see Table \ref{table:ext_if}). 
  Finally, 
  \item entries in \lex pointing to entities of only one ontology are not considered (see last two rows of \lex in Table \ref{table:ext_if}).
\end{inparaenum}
Note that a single entity label may lead to several entries in \lex, and
each entry in \lex points to one or many entities.


Each entry in \lex, after discarding entries pointing to only one ontology, is a source of candidate mappings. For
instance the example in Table \ref{table:ext_if} suggests that there is a
(potential) mapping $m=\mapping{\cem{\ontoOne \nsp:\nsp
Serous\_acinus}}{\cem{\ontoTwo \nsp:\nsp Liver\_acinus}}{\equiv}{c}$ since the entities 
$\cem{\ontoOne \nsp:\nsp Serous\_acinus}$ and $\cem{\ontoTwo \nsp:\nsp Liver\_acinus}$ are associated to
the same entry in \lex \emph{\{acinus\}}.
These mappings are not necessarily correct but will link lexically-related entities, that is, those entities
sharing at least one word among their labels (\eg ``acinus'').
Given a subset of entries of \lex (\ie $l \subseteq \lex$), the function 
$\funcalgo{Mappings}(l)=\M^{l}$ provides the set of mappings derived from $l$.  We refer to the
set of all (potential) mappings suggested by \lex (\ie $\funcalgo{Mappings}(\lex)$) as $\MLex$. Note that $\MLex$ represents a manageable subset of the Cartesian product between the entities of the input ontologies. 

Most of the state-of-the-art ontology matching systems rely, in one way or
another, on lexical similarity measures to either discover or validate
candidate mappings \cite{DBLP:journals/tkde/ShvaikoE13, DBLP:books/daglib/0032976}. Thus, mappings outside $\MLex$ will rarely be
discovered by standard matching systems. 

\vspace{-0.1cm}
\paragraph{Dealing with limited lexical overlapping.}
The construction of \lex, which is the basis of the methods
presented in this section, shares a limitation with state-of-the-art systems, that is, 
the input ontologies are lexically disparate or do not provide enough lexical information. In this case, the set of mapping $\MLex$ may be too small or even empty.
As a standard solution, if the ontologies have a small lexical overlapping or are in different
languages, \lex can be enriched with general-purpose lexicons (\eg
WordNet or the UMLS lexicon), more specialised background knowledge (\eg UMLS Metathesaurus) or with translated labels using online translation services like the ones provided by Google, IBM or Microsoft.

\subsection{Overlapping estimation}

The mappings in $\MLex$ can be used to extract an (over)estimation of the 
overlapping between the ontologies $\ontoOne$ and $\ontoTwo$. 


\begin{definition}[Extended overlapping matching task]
\label{def:extoverlapping}
Let $\MLex$ be the alignment computed from \lex for $\O_1$ and
$\O_2$, 
and $\funcalgo{Context}(\MLex, \O_1, \O_2)=\langle
\ontoOneP{\lex}, \ontoTwoP{\lex} \rangle$ the context of $\MLex$. We define
$\MTLex=\langle \ontoOneP{\lex}, \ontoTwoP{\lex} \rangle$ as the extended overlapping matching~task.
\end{definition}

$\MTLex=\langle \ontoOneP{\lex}, \ontoTwoP{\lex} \rangle$ can also be seen as the
result of reducing or dividing the task $\MT=\langle \ontoOne, \ontoTwo \rangle$
where only one matching subtask is given as output (\ie
$\PMT^1=\{\MTLex\}$).

\begin{hypothesis}
\label{prop:mlex}
If $\MT=\langle \ontoOne, \ontoTwo \rangle$ is a matching task, $\MS$
the mappings computed for $\MT$ by a lexical-based matching system, and
$\PMT^1=\{\MTLex\}$ the reduction of the matching task $\MT$ using the
notion of overlapping (over)estimation, then $\PMT^1$ covers (almost) all the
mappings in $\MS$, that is, $\funcalgo{CoverageRatio}(\PMT^1, \MS) \approx 1.0$.
\end{hypothesis}


Hypothesis \ref{prop:mlex} suggests that a matching system will unlikely
discover mappings with $\MT=\langle \ontoOne, \ontoTwo \rangle$ that cannot be
discovered with  $\PMT^1=\{\MTLex\}$. This
intuition is not only supported by the observation that most of the ontology
matching systems rely on lexical similarity, but also by the use of the notion
of context (see Definition~\ref{def:context1} and Definition
\ref{def:overlappingMT}) in the creation of the \emph{extended overlapping matching task}.



\subsection{Creation of matching subtasks from \lex}

Considering all entries in \lex 
(\ie one cluster) 
may lead to a
very large number of candidate mappings $\MLex$ and, as a consequence, to large overlapping modules
$\ontoOneP{\lex}$ and~$\ontoTwoP{\lex}$. These modules, although smaller than
$\ontoOne$ and $\ontoTwo$, can still be challenging for many ontology matching systems. A solution is to divide the entries in \lex in more than one cluster. 

\begin{definition}[Matching subtasks from \lex]
\label{def:partIF}
Let $\MT=\langle \ontoOne, \ontoTwo \rangle$ be a matching task, \lex the lexical index of the ontologies $\ontoOne$ and $\ontoTwo$, and $\{c_1,\ldots,c_n\}$ a set of $n$ clusters of entries in \lex. 
We denote the set of matching subtasks from \lex as
$\PMT^{n} = \{\MTLex_1,\ldots,\MTLex_n\}$ where each cluster $c_i$ leads to the matching subtask $\MTLex_i=\langle \ontoOneP{i},
\ontoTwoP{i} \rangle$, such that $\funcalgo{Mappings}(c_i) = \MLex_i$ is the set of mappings suggested by the \lex entries in $c_i$
and $\ontoOneP{i}$ and $\ontoTwoP{i}$ represent the context of $\MLex_i$ w.r.t. $\ontoOne$ and $\ontoTwo$.
\end{definition}

Since the matching subtasks in Definition \ref{def:partIF} also rely on \lex and the notion of context of the derived mappings from each cluster of entries in \lex, it is expected that the resulting matching subtasks in $\PMT^{n}$ will have a coverage similar to $\PMT^1$.


\begin{hypothesis}
\label{prop:coverage_lexical_part}
If $\MT=\langle \ontoOne, \ontoTwo \rangle$ is a matching task and $\MS$
the mappings computed for $\MT$ by a lexical-based matching system, then,
with independence of the clustering strategy of \lex and the
number of matching subtasks $n$, $\PMT^{n} =
\{\MTLex_1,\ldots,\MTLex_n\}$ will cover (almost) all the mappings in $\MS$
(\ie $\funcalgo{CoverageRatio}(\PMT^n, \MS) \approx 1.0$).
\end{hypothesis}

Intuitively each cluster of \lex will lead to a smaller set of
mappings $\MLex_i$ (with respect to $\MLex$) and to a smaller matching task $\MTLex_i$ (with respect to both $\MTLex$ and $\MT$) in terms of
search space. Hence $\funcalgo{SizeRatio}(\MTLex_i, \MT)$ will be smaller than $1.0$, as mentioned in Section
\ref{sect:quality}.
%
Reducing the search space
in each matching subtask $\MTLex_i$ has the potential of enabling 
the use of systems that
can not cope with the original matching task $\MT$ in a given time-frame or
with (limited) computational resources. The aggregation of ratios may be greater than
$1.0$ and will depend on the clustering strategy.

\begin{hypothesis}
\label{prop:new}
Given a matching task $\MT$ and an ontology matching system that fails to
complete $\MT$ under a set of given computational constraints, there exists 
a  
division of the matching task $\PMT^{n} = \{\MTLex_1,\ldots,\MTLex_n\}$ for which that system is
able to compute an alignment of the individual matching subtasks $\MTLex_1,\ldots,\MTLex_n$
under the same constraints.
\end{hypothesis}

Decreasing the search space may also improve the
performance of systems able to cope with $\MT$ in terms f-measure.

\begin{hypothesis}
\label{prop:performance}
If $\MT=\langle \ontoOne, \ontoTwo \rangle$ is a matching task, $\MS$
the mappings computed for $\MT$ by a state-of-the-art matching system and
$F$ the f-measure of $\MS$ w.r.t. a given reference alignment $\MRA$, then the
set of mappings $\MS_p=\MS_1\cup\ldots\cup\MS_n$ computed by the same system
over the matching subtasks in $\PMT^{n}=\{\MTLex_1,\ldots,\MTLex_n\}$ leads
 to an f-measure $F'$ such that $F' \ge F$.
\end{hypothesis}

Hypothesis \ref{prop:performance} is based on the observation that systems in
the OAEI \emph{largebio} track~\cite{DBLP:conf/semweb/AchichiCDEFFFFH17} show a better performance when, instead of the original matching task (\eg whole FMA and NCI ontologies), they are given the overlapping matching task for the reference alignments (as in Definition \ref{def:overlappingMT}).


\subsection{Clustering strategies}

We have implemented two clustering strategies  which we refer to as:
\emph{naive} and \emph{neural embedding}. 
Both strategies receive as input 
the index \lex and the number of desired clusters $n$, and provide as output a set of  clusters $\{c_1,\ldots,c_n\}$ from \lex. As in Definition \ref{def:partIF}, these cluster lead to the set of matching subtasks $\PMT^{n} =
\{\MTLex_1,\ldots,\MTLex_n\}$.

The choice of strategy, according to
Hypothesis~\ref{prop:coverage_lexical_part}, will not have an impact on the 
coverage; but it may influence the size of the matching subtasks. Note that,
neither of the strategies aim at computing optimal clusters of the entries in 
\lex, but clusters that can be efficiently computed. 

\paragraph{Naive strategy.} This strategy implements a very simple algorithm
that randomly splits the entries in \lex into a given number of
clusters of the same size. 
The matching tasks resulting from this strategy are expected to have
a high overlapping as different entries in \lex leading to similar set of mappings may
fall into different clusters. Although the overlapping of matching subtasks 
will impact the global search space, it is still expected to be smaller than in the original matching task.

\paragraph{Neural embedding strategy.} This strategy aims at 
identifying more accurate clusters, leading to matching tasks with less overlapping, and thus, reducing
the global size of the computed division of the matching task $\PMT^{n}$.
%
%
It relies on \starspace toolkit\footnote{\starspace: \url{https://github.com/facebookresearch/StarSpace}} and its neural embedding model~\cite{wu_2017}, 
which aims at learning \emph{entity embeddings}. Each entity\footnote{Note that in the context of neural embedding models the term entity refers to objects of different kind, \eg a word, a sentence, a document or even an ontology entity.} is described by a finite set of discrete \emph{features} (bag-of-features).
The model is trained by assigning a $d$-dimensional vector to each
of the discrete features in the set that we want to embed directly. Ultimately,
the look-up matrix (the matrix of embeddings - latent vectors) is learned by
minimizing the loss function in Equation \ref{eq:loss}. %
\begin{equation}\label{eq:loss}
\sum_{(k, v) \in E^+, v^- \in E^-}
L^{batch} (sim(k, v), sim(k, v_1^-), \ldots, sim(k, v_k^-))
\end{equation}
%
In this loss function, we need to indicate the generator of positive entry
pairs $(k, v) \in E^+$ --  in our setting those are \emph{key-value} pairs from \lex -- and the generator of negative entries
$(k, v^-) \in E^-$ (the so-called \emph{negative examples})
-- in our setting, 
the pairs $(k, v^-)$ that do not appear in \lex. The similarity function
$sim$ is task-dependent and should operate on $d$-dimensional vector
representations of the entities, in our case we use the standard Euclidean dot
product. The aforementioned neural embedding model corresponds to the
\texttt{TagSpace} training setting of \starspace (see~\cite{wu_2017} for more details). 
Applied to the lexical index \lex, the neural embedding model would learn vector
representations for the individual words in the index keys,
and for the individual entity identifiers in the index values.
Since an index key 
is a set of words (see Table \ref{table:ext_if}), we
use the \textit{mean vector} representation of the vectors associated to each
word
(in principle other \textit{aggregated}
representation could be applied).
Based on these \textit{aggregated} neural embeddings we then perform standard clustering with the
K-means algorithm. 
%

\vspace{-0.075cm}
\begin{hypothesis}
\label{prop:sizetask}
There exists a number of clusters or matching subtasks `$n$' for which the
clustering strategies can compute  $\PMT^{n} = \{\MTLex_1,\ldots,\MTLex_n\}$ for
a given matching task $\MT$ such that $\funcalgo{SizeRatio}(\PMT^{n}, \MT) <
1.0$.
\end{hypothesis}

\vspace{-0.075cm}
Hypothesis \ref{prop:sizetask} suggests that there exists a division 
$\PMT^{n}$ of $\MT$ such that the size (or search space) of $\PMT^{n}$ is
smaller than $\MT$, and $\PMT^{n}$ can be computed by the proposed naive
and neural embedding strategies.

\section{Evaluation}
\label{sec:eval}

\begin{table}[t!]
\caption{Matching tasks. AMA: Adult
Mouse  Anatomy. DOID: Human Disease Ontology. FMA: Foundational Model
of Anatomy. HPO: Human Phenotype Ontology.  MP: Mammalian Phenotype. NCI: National
Cancer Institute. NCIA: Anatomy fragment of NCI. ORDO: Orphanet Rare
Disease Ontology. SNOMED: Systematized Nomenclature of Medicine -- Clinical
Terms. Phenotype ontologies downloaded from BioPortal.}\label{table:tasks}
\begin{tabular}{|c|c|c||c|c|c|}
\hline
\textbf{OAEI track} & \textbf{Source of $\MRA$} & \textbf{Task}  &
\textbf{Ontology} & \textbf{Version} & \textbf{Size (classes)} \\\hline\hline

\multirow{2}{*}{Anatomy} &  \multirow{2}{*}{Manually
created} & \multirow{2}{*}{AMA-NCIA} & AMA & v.2007 & 2,744 \\
& & & NCIA & v.2007 & 3,304 \\\hline\hline

\multirow{3}{*}{Largebio} & \multirow{3}{*}{UMLS-Metathesaurus} & FMA-NCI & FMA
& v.2.0 & 78,989 \\
& & ~FMA-SNOMED~ & NCI & v.08.05d &  66,724 \\
& & SNOMED-NCI & ~SNOMED~ &  v.2009 &  306,591\\\hline\hline

\multirow{4}{*}{Phenotype} & 
\multirow{4}{*}{\vtop{\hbox{\strut Consensus alignment}\hbox{\strut (vote=2)
\cite{phenotype2017}}}}

& \multirow{2}{*}{HPO-MP} & HPO &
~v.2016-BP~ & 11,786 \\
& & & MP & v.2016-BP & 11,721 \\
& &  \multirow{2}{*}{DOID-ORDO} & DOID & v.2016-BP & 9,248 \\
& & & ORDO & v.2016-BP & 12,936 \\\hline
\end{tabular}
\end{table}

In this section we aim at providing empirical evidence to support the Hypothesis
\ref{prop:mlex}-\ref{prop:sizetask} introduced in Section
\ref{sec:methods}.
We rely on the datasets of the Ontology Alignment Evaluation Initiative (OAEI) 
\cite{DBLP:conf/semweb/AchichiCDEFFFFH17}, more specifically, on the matching
tasks provided in the \emph{anatomy},
\emph{largebio}
and \emph{phenotype}
tracks.
Table \ref{table:tasks} provides an overview of these OAEI tasks and the related ontologies.
%

The methods have been implemented in Java\footnote{Java codes:
\url{https://github.com/ernestojimenezruiz/logmap-matcher}} and Python\footnote{Python codes: \url{https://github.com/plumdeq/neuro-onto-part}} (neural embedding
strategy), tested on a Ubuntu Laptop with an Intel Core i7-4600U CPU@2.10GHz (4 cores) and allocating up to 15 Gb of RAM. 
Datasets, evaluation results, logs and other supporting resources are
available in the \emph{Zenodo} repository~\cite{zenodo_material_iswc}.

%
%

We have performed the following experiments, which we describe in detail in the following sections:%
\begin{itemize}[noitemsep,topsep=0pt]
  \item We have computed the extended  overlapping matching task (\ie $\PMT^1$) for each of
  the matching tasks as in Definition \ref{def:extoverlapping} and calculated the
  coverage with respect to the available reference alignments (Section~\ref{sect:evalOverlapping}).
  \item We have applied the \emph{naive} and \emph{neural
  embedding}\footnote{Please refer to \cite{zenodo_material_iswc} for
  information about the used \starspace input parameters.} strategies to compute divisions $\PMT^n$ of the matching tasks and
  evaluated their adequacy with respect to coverage and size
  (Section~\ref{sect:evalPartitioning}).
  \item We have evaluated the performance of a selection of OAEI matching
  systems over the computed matching subtasks and compared with their original results
  (if any) in the OAEI campaigns (Section~\ref{sect:evalSystems}).  
\end{itemize}
%
%

\subsection{Coverage of the extended overlapping matching task}
\label{sect:evalOverlapping}

\begin{table}[t!]
\caption{Coverage results for $\PMT^1$}\label{tab:part1}
\centering

\begin{tabular}{|c|c||c|c|c|c|c|}

\hline
\multirow{2}{*}{\textbf{Task}} &
\multirow{2}{*}{~~\textbf{$\lvert$\lexN$\rvert$}~~} &
\multicolumn{5}{c|}{\boldmath{$\PMT^1$} \textbf{statistics}} \\\cline{3-7}

& & ~~~\boldmath{$\lvert\ontoOneP{1}\rvert$}~~~ &
~~~\boldmath{$\lvert\ontoTwoP{1}\rvert$}~~~ & ~~~$\bm{\funcalgo{SizeRatio}}$~~~
& $\bm{\funcalgo{CoverageRatio}}$ & ~~\textbf{time (s)}~~ \\\hline\hline

AMA-NCIA & 4,048 & 2,518 & 2,841 & 0.784 & 0.982 & 0.55 \\\hline
FMA-NCI & 11,507 & 33,744 & 35,409 & 0.226 & 0.994 & 10.3 \\\hline
FMA-SNOMED & 29,677 & 55,469 & 119,488 & 0.273 & 0.982 & 28.8 \\\hline
SNOMED-NCI & 45,940 & 190,911 & 56,076 & 0.521 & 0.968 & 28.2 \\\hline
HPO-MP & 10,514 & 8,165 & 10,041 & 0.589 & 0.995 & 1.93 \\\hline
DOID-ORDO & 13,375 & 7,166 & 10,741 & 0.637 & 0.999 & 2.81 \\\hline

\end{tabular}
\end{table}

We have evaluated the coverage
of $\PMT^1=\{\MTLex\}$ computed for each of the matching tasks in
Table~\ref{table:tasks} with respect to the available reference alignments. Table~\ref{tab:part1} summarizes the obtained results.
The second column of the table gives the number of entries in \lex,
while the last column represents the time to compute \lex, the derived mappings $\MLex$ and the
context of $\MLex$ (\ie the overlapping matching~task).
The obtained coverage (ratio) values range from $0.967$ to $0.999$, which
strongly supports our intuitions behind Hypothesis \ref{prop:mlex}. Furthermore,
since we have calculated the coverage with respect to the reference
alignments instead of system mappings (\ie $\funcalgo{CoverageRatio}(\PMT^1,
\MRA)$), the results also suggest that the information loss with respect to
system-generated alignments will be minimal. At the same time the size (ratio)
of the matching tasks is significantly reduced for the largest matching tasks. For example, for the FMA-NCI case, the resulting
task size has been reduced to $27.3\%$  of the original task size. 
The achieved high coverage in combination with the reduction of the search space
and the small computation times provide empirical evidence of the
suitability of \lex to reduce the alignment task at hand.



\subsection{Adequacy of the clustering strategies}
\label{sect:evalPartitioning}

We have evaluated the adequacy of the clustering strategies in terms of
coverage (as in Equation \ref{eq:covratio}) and size (as in Equation
\ref{eq:sizeeratio}) of the resulting division $\PMT^n$ of the matching task.
We have compared the two strategies for different number of clusters or resulting matching subtasks
$n \in \{2, 5, 10, 20, 50, 100, 200\}$.
 For the naive strategy, as a
random split of \lex is performed, we run 10 experiments for each
of the values of $n$
to evaluate the effect of different random selections. The
variations in the size of the obtained matching tasks was
negligible.\footnote{Details about matching task sizes and standard deviations can be found in
\cite{zenodo_material_iswc}.} The results reported for the naive strategy
represent the average of the 10 experiments.

\begin{figure}[t]
    \centering
    \begin{subfigure}[b]{0.495\textwidth}
        \centering
        \includegraphics[width=\textwidth]{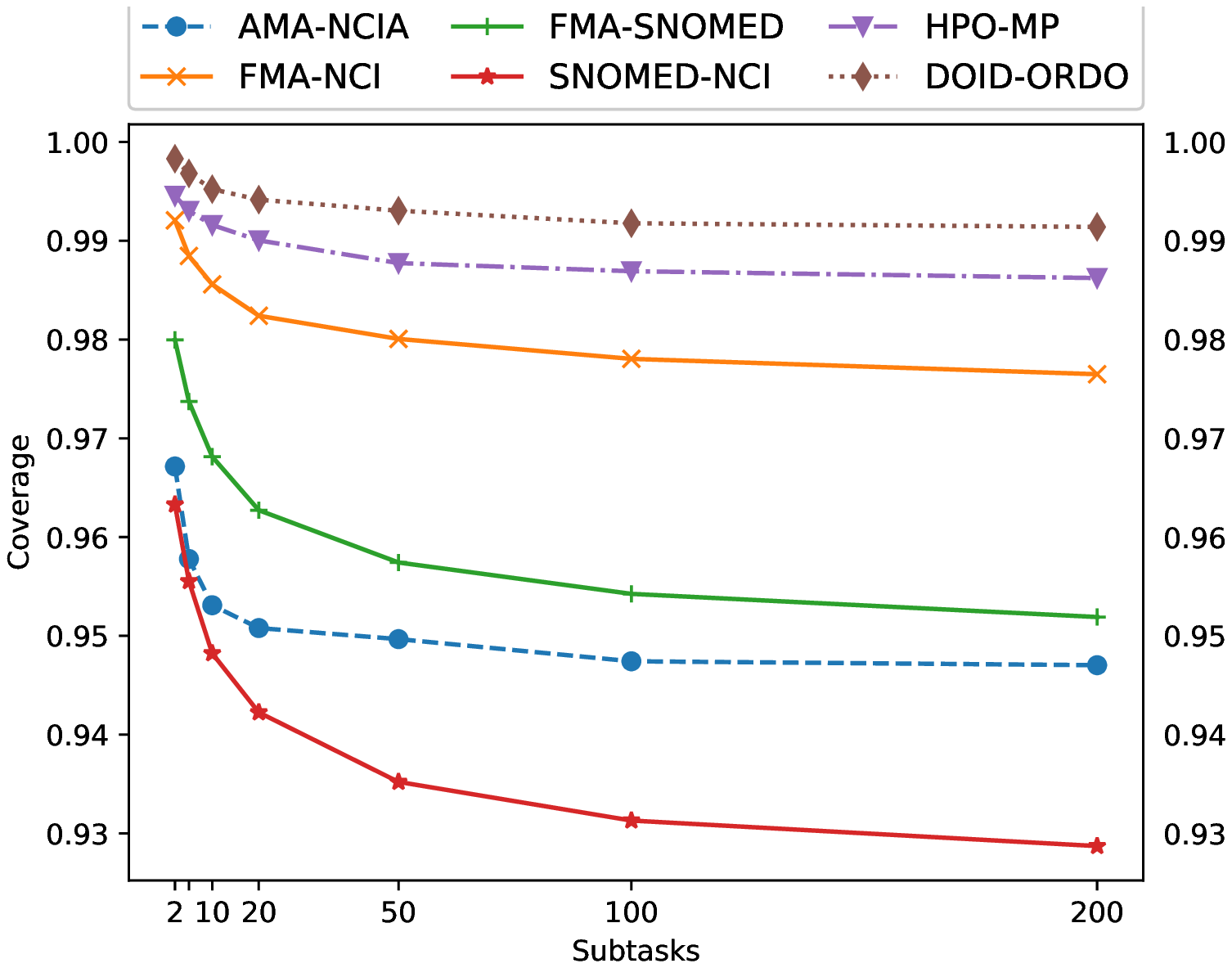}\\[-1ex]
        \caption{Naive strategy}
        \label{fig:coverageN}
    \end{subfigure}
    \begin{subfigure}[b]{0.495\textwidth}
        \centering
        \includegraphics[width=\textwidth]{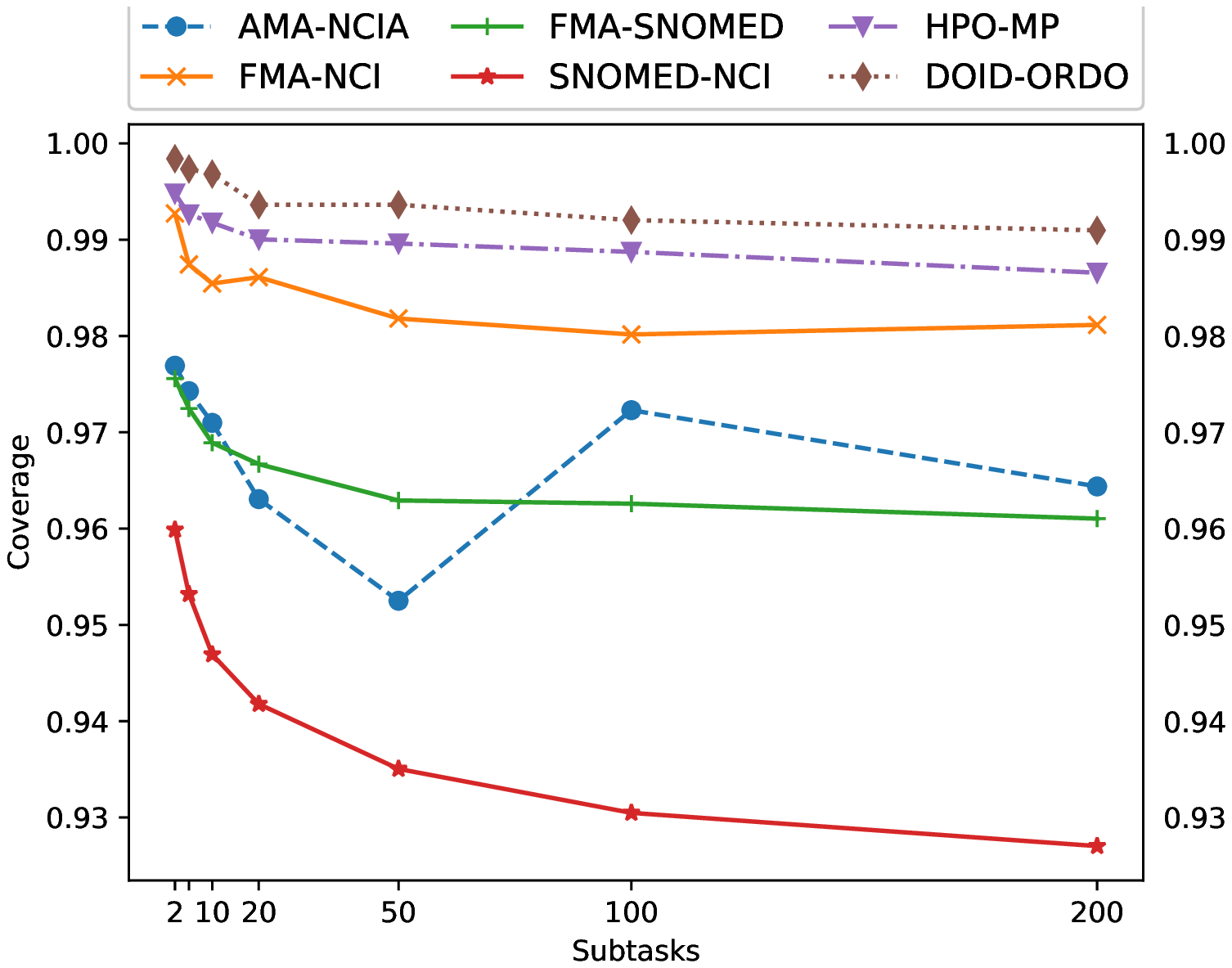}\\[-1ex]
        \caption{Neural embedding strategy}
        \label{fig:coverageA}
    \end{subfigure}
    \caption{$\funcalgo{CoverageRatio}$ of $\PMT^n$ with respect to the number
    of matching subtasks $n$.}
    \label{fig:coverages}
\end{figure}

\begin{figure}[t]
    \centering
    \begin{subfigure}[b]{0.495\textwidth}
        \centering
        \includegraphics[width=\textwidth]{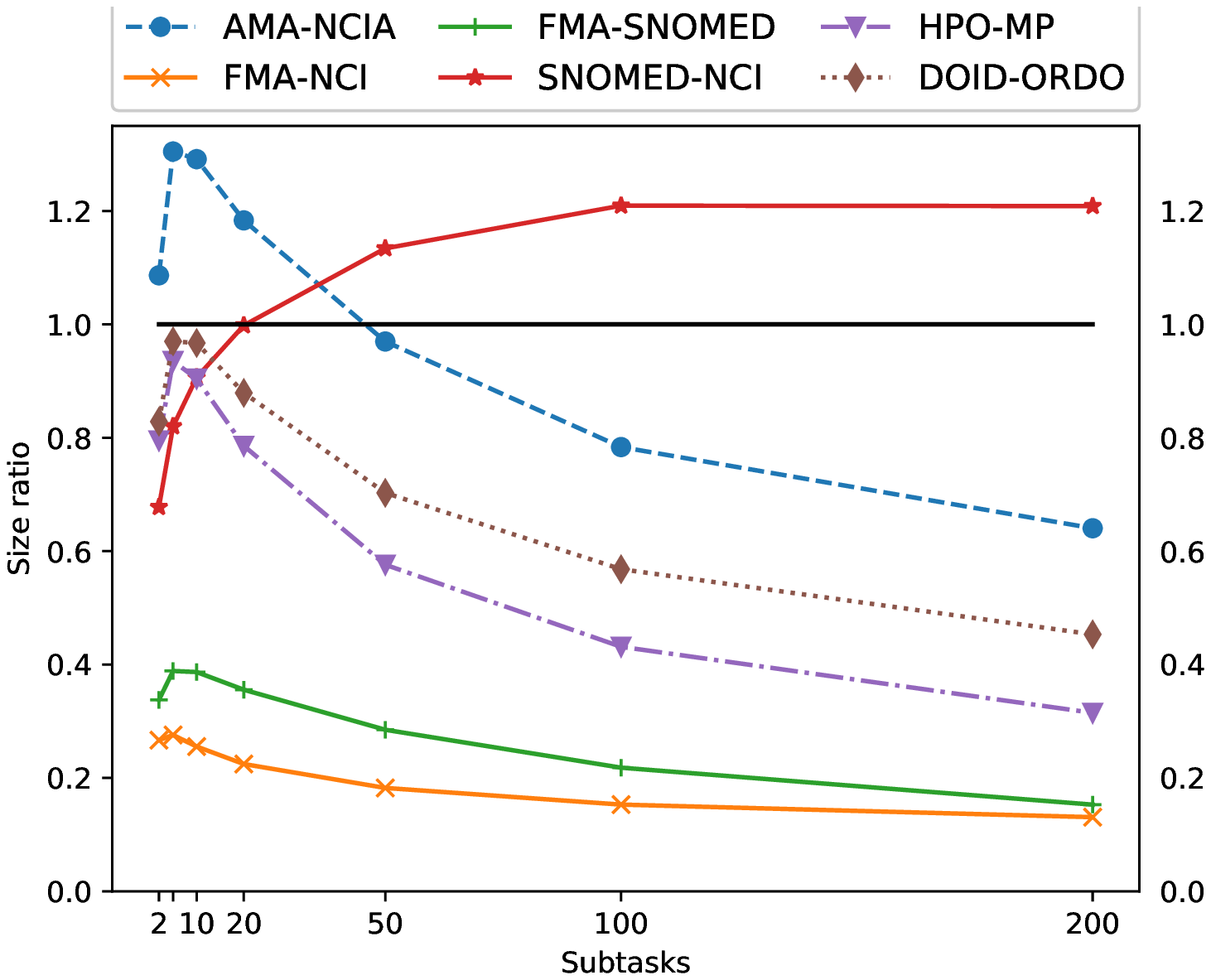}\\[-1ex]
        \caption{Naive strategy}
        \label{fig:ratioN}
    \end{subfigure}
    \begin{subfigure}[b]{0.495\textwidth}
        \centering
        \includegraphics[width=\textwidth]{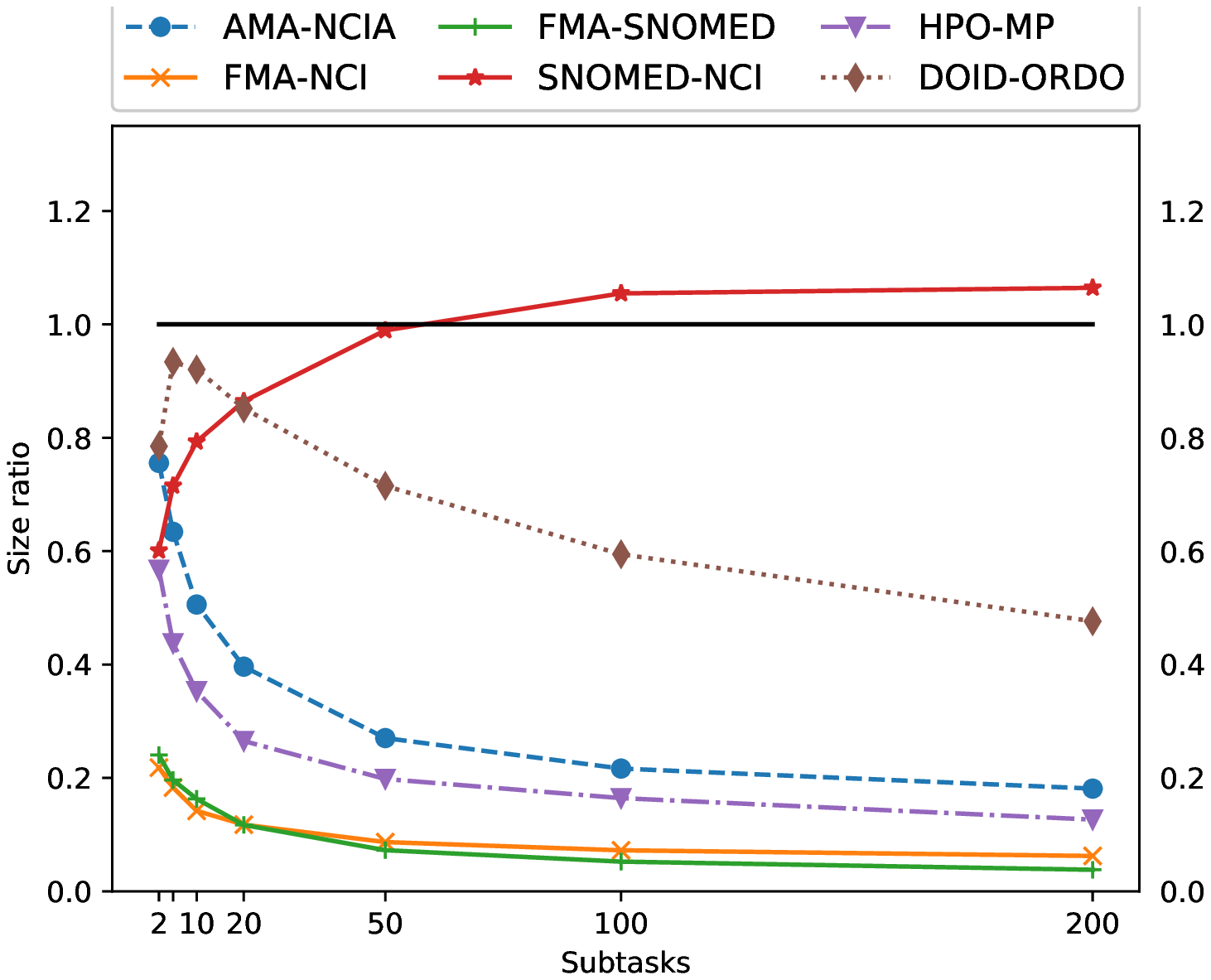}\\[-1ex]
        \caption{Neural embedding strategy}
        \label{fig:ratioA}
    \end{subfigure}
    \caption{$\funcalgo{SizeRatio}$ of $\PMT^n$ with respect to the number of
    matching subtasks $n$.}
    \label{fig:ratios}
\end{figure}

\vspace{-0.25cm}
\paragraph{Coverage ratio.}
Figure \ref{fig:coverages} shows the coverage of the different divisions
$\PMT^n$ of the matching task for the naive (left) and neural embedding (right) strategies. As in the
case of $\PMT^1=\{\MT^{Lex}\}$ the coverage ratio is very good, being
$0.927$ in the worst case ($n=200$ in SNOMED-NCI) and $0.99$ in the best case
($n=2$ in FMA-NCI). This means that, in the worst case, almost $93\%$ of the
available reference mappings are \emph{covered} by the matching subtasks
in $\PMT^n$.
The differences in terms of coverage between the naive and neural embedding strategies are
minimal, with the neural embedding strategy providing slightly better results
on~average. These results reinforce Hypothesis \ref{prop:coverage_lexical_part}
as the coverage with respect to system-generated mappings is expected to be even
better.

\vspace{-0.15cm}
\paragraph{Size ratio.} The results in terms of the size (\ie search space) of
the selected divisions $\PMT^n$ are
presented in Figure~\ref{fig:ratios} for the naive (left) and neural embedding
(right) strategies.
The results with the neural embedding strategy are extremely positive, while the results
of the naive strategy, although slightly worse as expected, are
surprisingly very competitive.
Both strategies improve the
search space with respect to the original $\MT$ for all cases with the exception
of the naive strategy in the AMA-NCIA case with $n<50$, and the SNOMED-NCI case
with $n>20$, which 
validates Hypothesis
\ref{prop:sizetask}.
SNOMED-NCI confirms to be the hardest case in the
\emph{largebio} track. Here the size ratio increases with the number of matching subtasks $n$ and gets stable with~$n>100$.

\begin{figure}[t]
    \centering
    \begin{subfigure}[b]{0.495\textwidth}
        \centering
        \includegraphics[width=\textwidth]{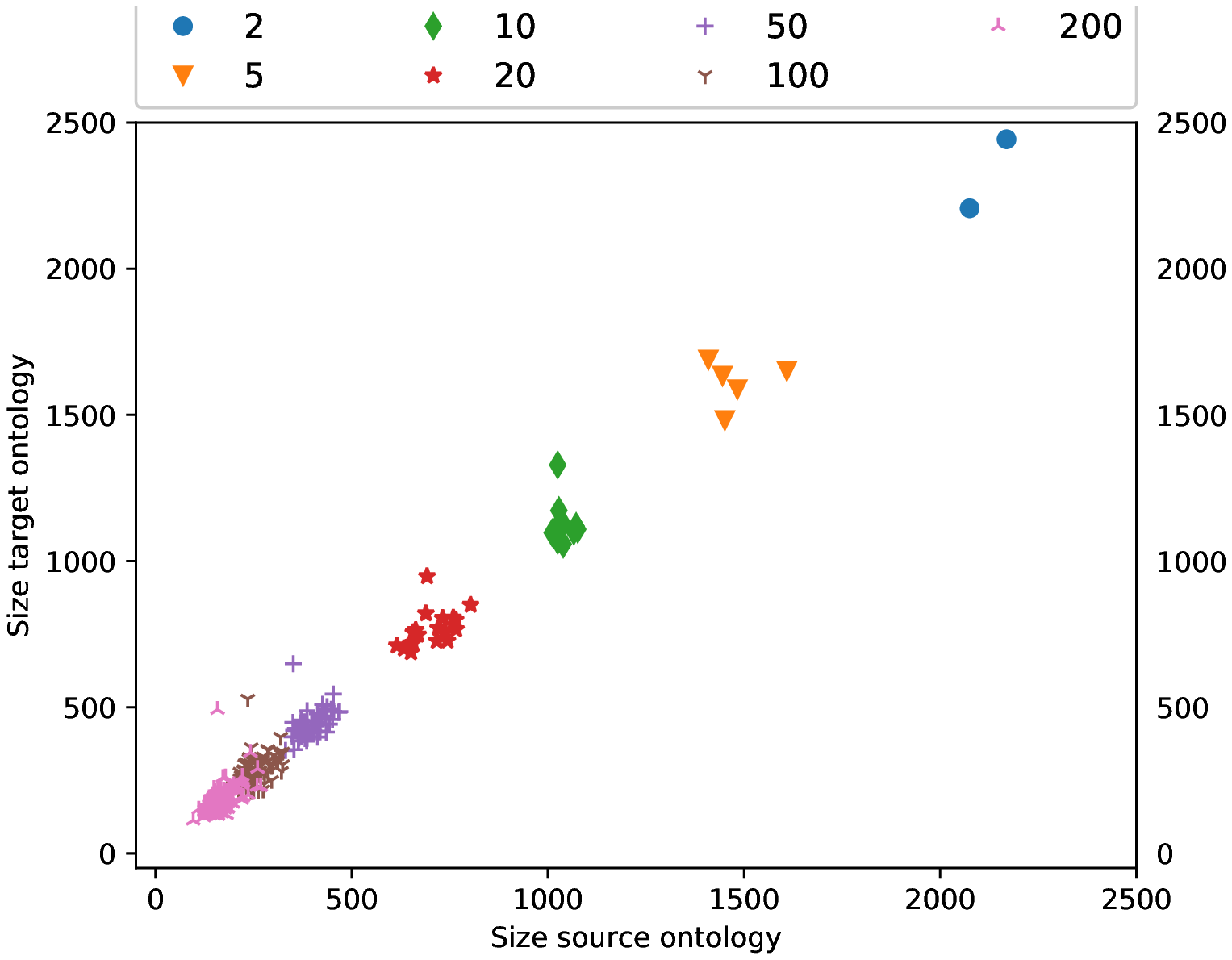}\\[-1ex]
        \caption{Naive strategy}
        \label{fig:mouse-naive}
    \end{subfigure}
    \begin{subfigure}[b]{0.495\textwidth}
        \centering
        \includegraphics[width=\textwidth]{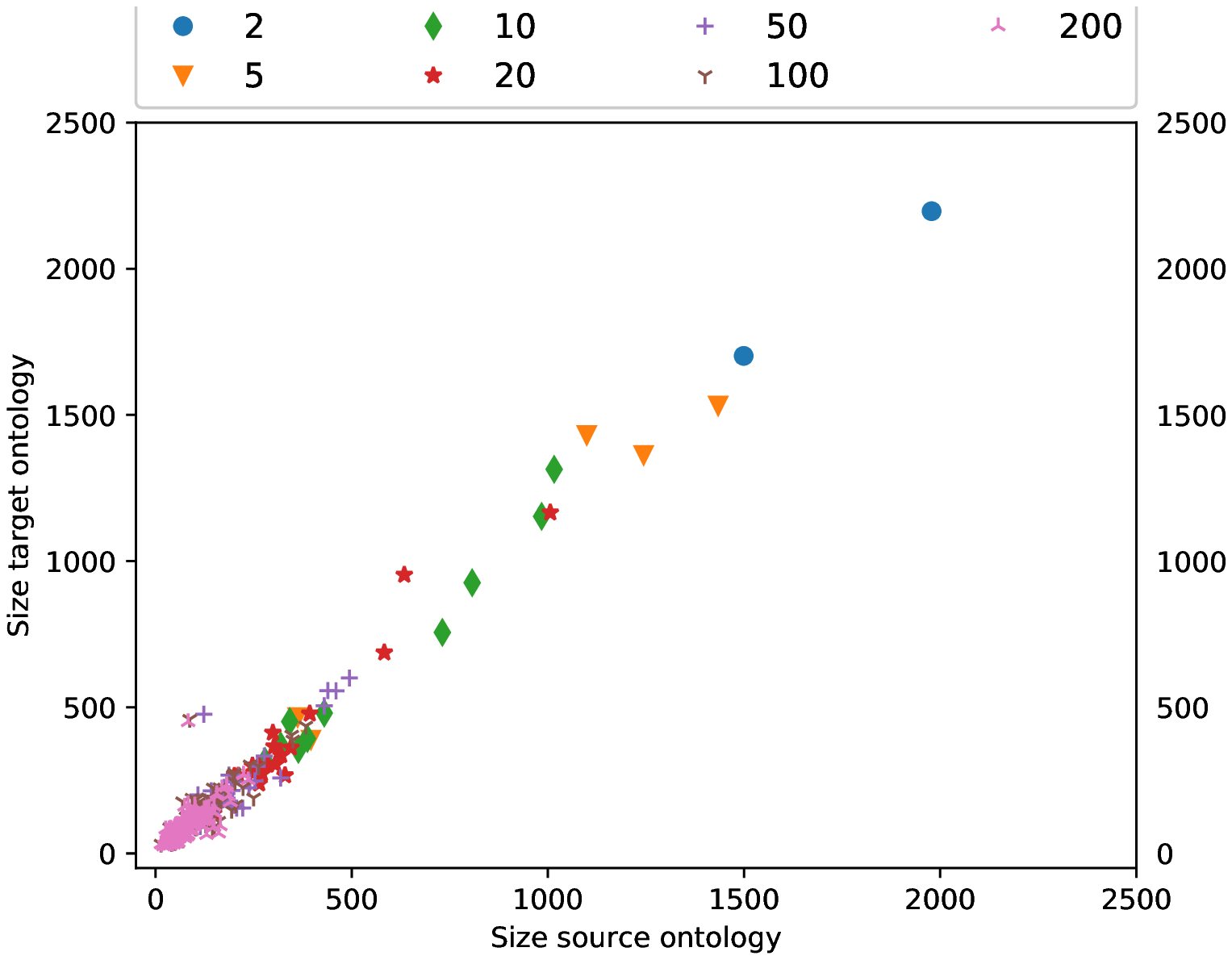}\\[-1ex]
        \caption{Neural embedding strategy}
        \label{fig:mouse-advanced}
    \end{subfigure}
    \caption{Source and target module sizes in the computed subtasks  for
    AMA-NCIA.}
    \label{fig:clouds1}
\end{figure}
    
\begin{figure}[t]
    \centering
    \begin{subfigure}[b]{0.495\textwidth}
        \centering
        \includegraphics[width=\textwidth]{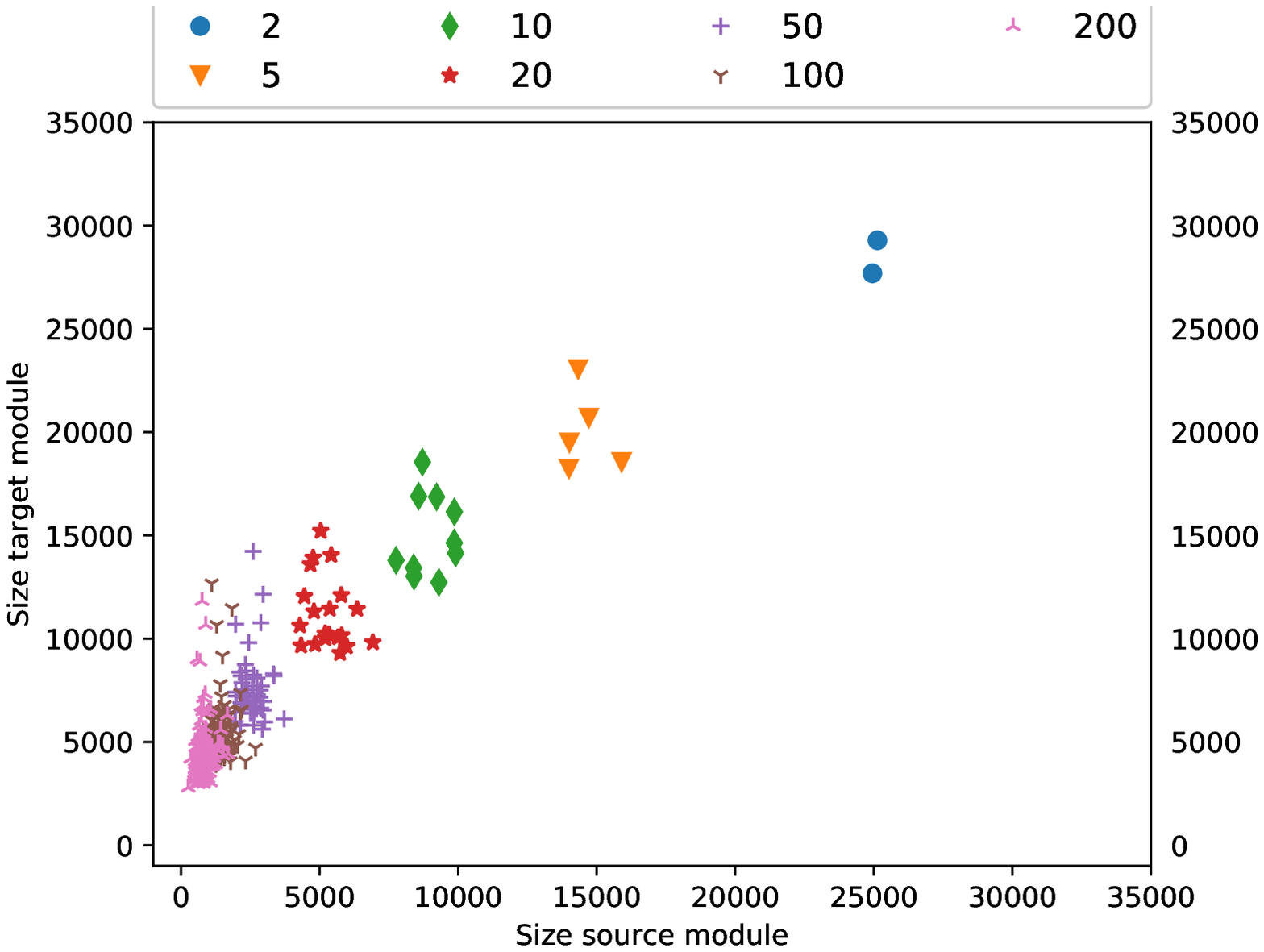}\\[-1ex]
        \caption{Naive strategy}
        \label{fig:f2n-naive}
    \end{subfigure}
    \begin{subfigure}[b]{0.495\textwidth}
        \centering
        \includegraphics[width=\textwidth]{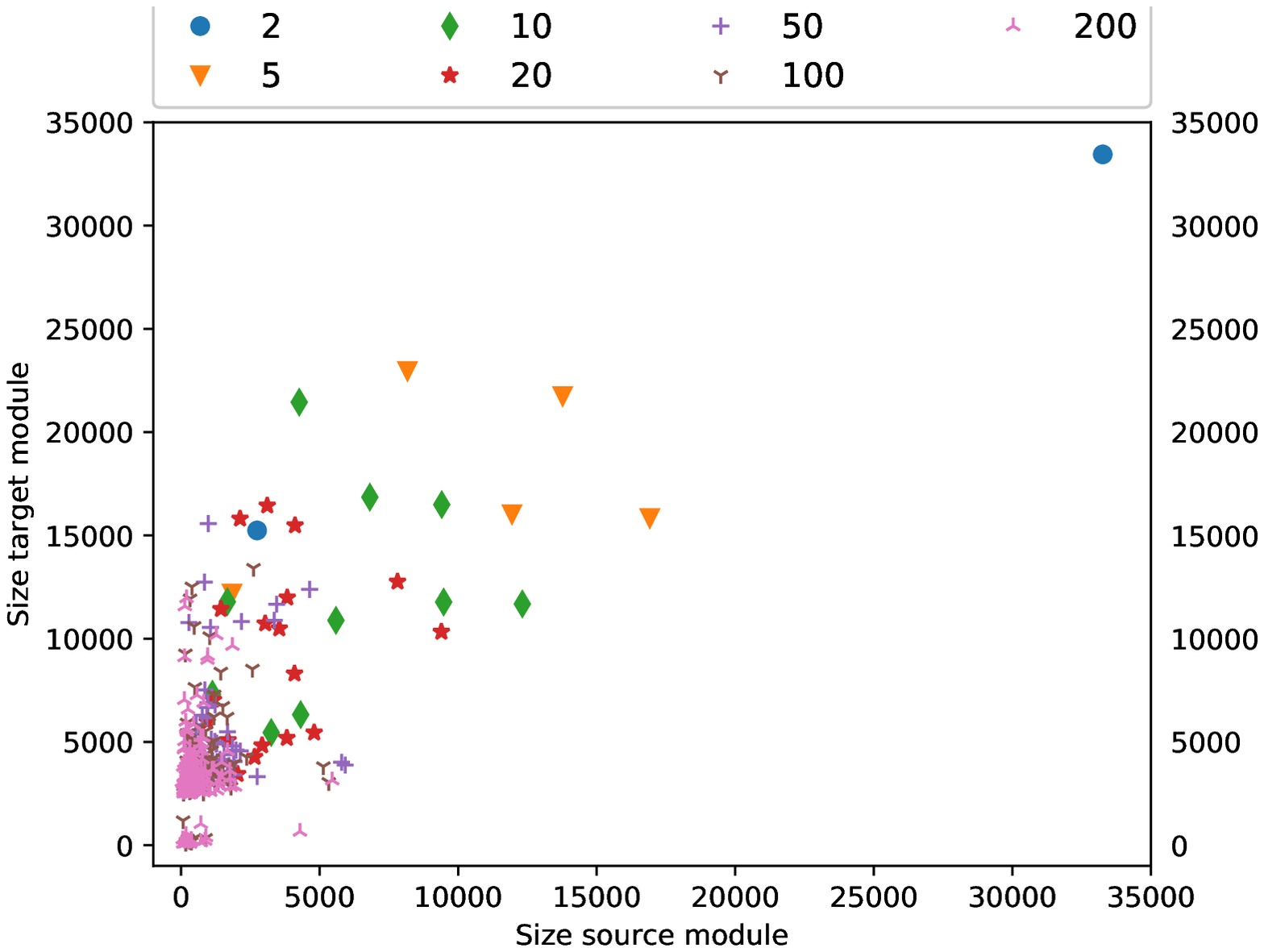}\\[-1ex]
        \caption{Neural embedding strategy}
        \label{fig:f2n-advanced}
    \end{subfigure}
    \caption{Source and target module sizes in the computed subtasks for FMA-NCI.}
    \label{fig:clouds2}
\end{figure}

\vspace{-0.15cm}
\paragraph{Size of the source and target modules.} The scatter plots in Figures
\ref{fig:clouds1} and \ref{fig:clouds2} visualize the size of the source
modules against the size of the target modules for the matching tasks in each division $\PMT^n$. For
instance, the (orange) triangles represent points $\big(\lvert Sig(\ontoOne^{i})
\rvert, \lvert Sig(\ontoTwo^{i}) \rvert \big)$ being
$\ontoOne^{i}$ and $\ontoTwo^{i}$ the source and target modules (with $i$=$1$,\ldots,$5$)
in the matching subtasks of~$\PMT^5$.
Figure
\ref{fig:clouds1} shows the plots for the AMA-NCIA case while Figure
\ref{fig:clouds2} for the FMA-NCI case, using the naive (left) and neural embedding
(right) strategies. The naive strategy leads to rather balanced an similar tasks (note differentiated 
cloud of points) 
for each division $\PMT^n$ for both cases. The neural embedding strategy has
more variability in the size of the tasks within a given division $\PMT^n$. In
the FMA-NCI case the tasks generated by the neural embedding strategy are also less
balanced and the target module tends to be larger than the source module.
Nonetheless, on average, the (aggregated) size of the matching tasks in the
neural embedding strategy are significantly reduced as shown in Figure~\ref{fig:ratios}.

\vspace{-0.15cm}
\paragraph{Computation times.} The time to compute the divisions of the matching task is
tied to the number of locality modules to extract, which can be
computed in polynomial time relative to the size of
the input ontology\cite{DBLP:journals/jair/GrauHKS08}.
The creation of \lex does not add an important overhead, while the training of
the neural embedding model in the advance strategy ranges from 21s in AMA-NCI to 224s in SNOMED-NCI.
Overall, for example, the required time to compute the division with 50 matching subtasks
ranges from 2s in AMA-NCIA to 413s in SNOMED-NCI with the naive strategy, and from 24s
(AMA-NCIA) to 647s (SNOMED-NCI) with the neural embedding strategy. Complete list of
relevant times can be obtained from \cite{zenodo_material_iswc}.

\subsection{Evaluation of OAEI systems}
\label{sect:evalSystems}

\begin{table}[t!]
\caption{Evaluation of systems that failed to complete OAEI
tasks in the 2015-2017 campaigns. (*) GMap was tested allocating 8Gb of
memory. Time reported in hours~(h).}\label{table:tools}
\centering
\begin{tabular}{|c|c|c|c||c|c|c|c||c|c|c|c|}
\hline

\multirow{2}{*}{\textbf{Tool}} & \multirow{2}{*}{\textbf{Task}} & 
\multirow{2}{*}{\textbf{Year}}
& 
\textbf{Matching} 
& \multicolumn{4}{c||}{\textbf{Naive
strategy}} & \multicolumn{4}{c|}{\textbf{Neural embedding
strategy}}\\\cline{5-12}

& & & \textbf{subtasks} &
~\textbf{P}~ &
~\textbf{R}~ &  ~\textbf{F}~ &  ~\textbf{t (h)~} &
~\textbf{P}~ &
~\textbf{R}~ &  ~\textbf{F}~ &  ~\textbf{t (h)~}
\\\hline\hline

\multirow{2}{*}{GMap (*)} & \multirow{2}{*}{Anatomy} & \multirow{2}{*}{2015} 
& 5 & 0.87 & 0.81 & 0.84 & 1.3 & 0.88 & 0.82 & 0.85 & 0.7\\
& & & 10 & 0.85 & 0.81 & 0.83 &	1.7 & 0.86 & 0.82 & 0.84 & 0.8
\\\hline\hline

\multirow{2}{*}{Mamba} & \multirow{2}{*}{Anatomy} & \multirow{2}{*}{2015} & 
20 & ~0.88~ & ~0.63~ & ~0.73~ & 2.3 & ~0.89~ & ~0.62~ & ~0.73~ & 1.0  \\
& & & 50 & 0.88 & 0.62 & 0.73 & 2.4 & 0.89 & 0.62 & 0.73 & 1.0
\\\hline\hline

\multirow{2}{*}{FCA-Map} & \multirow{2}{*}{FMA-NCI} & \multirow{2}{*}{2016} 
& 20 & 0.56 & 0.90 & 0.72 & 4.4 &  0.62 & 0.90 & 0.73 & 3.1 \\
& & & 50 & 0.58 & 0.90 & 0.70 &	4.1 & 0.60 & 0.90 & 0.72 & 3.0 \\\hline\hline

\multirow{2}{*}{KEPLER} & \multirow{2}{*}{FMA-NCI} & \multirow{2}{*}{2017} 
& 20 & 0.45 & 0.82 & 0.58 & 8.9 & 0.48 & 0.80 &	0.60 & 4.3\\
& & & 50 & 0.42 & 0.83 & 0.56 & 6.9 & 0.46 & 0.80 & 0.59 & 3.8
\\\hline\hline

\multirow{2}{*}{POMap} & \multirow{2}{*}{FMA-NCI} & \multirow{2}{*}{2017} 
& 20 & 0.54 & 0.83 & 0.66 &	11.9 & 0.56 & 0.79 & 0.66 & 5.7\\ 
& & & 50 & 0.55 & 0.83 & 0.66 & 8.8 & 0.57 & 0.79 & 0.66 & 4.1 \\\hline

 
\end{tabular}

\end{table}

In this section we support Hypothesis \ref{prop:new} by showing that
the division of the alignment task enables systems that, given
some computational constraints, were unable to complete an OAEI task. 
We have selected the following five systems
from the latest OAEI campaigns:\footnote{Other systems were also considered but they threw an exception during execution.}
Mamba, GMap, FCA-Map, KEPLER, and POMap.
Mamba and GMap failed to
complete the OAEI 2015 Anatomy track \cite{oaei2015} with 8Gb of allocated
memory, while FCA-Map, KEPLER and POMap could not complete the largest tasks in the \emph{largebio} track within a 12 hours
time-frame (with 16Gb of allocated memory) \cite{oaei2016,
DBLP:conf/semweb/AchichiCDEFFFFH17}.\footnote{In a preliminary evaluation round a 4 hours time-frame was given, which was later extended.} Note
that GMap and Mamba were also tested in the OAEI 2015 with 14Gb of memory. This
new setting allowed GMap to complete the task \cite{oaei2015}.

Table \ref{table:tools} shows the obtained results in terms of
computation times, precision, recall and f-measure over different
divisions $\PMT^n$ computed by the naive and neural embedding strategies. 
For example, Mamba was run over divisions
with 20 and 50 matching subtasks (\ie $n \in \{20, 50\}$).~Note that GMap was tested allocating only 8Gb of
memory as with this constraint it could not complete the task in the
OAEI 2015. The results can be summarized as follows:
\begin{enumerate}[\it i)]
  \item The computation times are encouraging since the (independent) matching
  tasks have been run sequentially without any type of parallelization.
  \item Times also include loading the ontologies from disk for
  each matching task. This step could be avoided if subtasks are
  directly provided by the presented framework.
  \item We did not perform an exhaustive analysis, but memory consumption
  was lower than 8Gb in all tests; thus, systems like GMap could run
  under limited resources.
  \item The increase of number of matching subtasks is beneficial for FCA-Map, KEPLER and
  POMap in terms of computation times. However, this is not the case for Mamba and GMap.
\item The division generated by the neural embedding strategy 
lead to smaller computation times than the naive
strategy counterparts, as expected from Figure~\ref{fig:ratios}. 
\item The f-measure is slightly reduced as the size of $n$ increases. This result does not support our intuitions behind Hypothesis \ref{prop:performance}.
\end{enumerate}

\vspace{-0.05cm}
\paragraph{Comparison with OAEI results.} There are \emph{baseline} results in
the OAEI for the selected systems \cite{oaei2015, oaei2016,
DBLP:conf/semweb/AchichiCDEFFFFH17}, with the exception of Mamba where the results are novel for the \emph{anatomy} track.
As mentioned before, GMap, if 14Gb were allocated, was able to complete the
\emph{anatomy} task and obtained an f-measure of $0.861$. KEPLER, POMap and FCA-Map completed
the OAEI task involving small fragments of FMA-NCI with an
f-measure of $0.891$, $0.861$ and $0.935$, respectively. The f-measure using the
divisions of the matching task is slightly lower for GMap, which once more, does not support our
Hypothesis \ref{prop:performance}. The results are much lower for the cases of
KEPLER, POMap and FCA-Map, but they cannot be fully 
comparable as systems
typically reduce their performance when dealing with the whole \emph{largebio}
ontologies \cite{DBLP:conf/semweb/AchichiCDEFFFFH17}. 
The authors of FCA-Map have also recently reported results for an improved
version of FCA-Map. They completed the FMA-NCI task in near 7 hours, with a
precision of $0.41$, a recall of $0.87$ and a f-measure of $0.56$. The results
obtained with $\PMT^{20}$ and $\PMT^{50}$ are thus very positive, since 
both strategies lead to much better numbers in terms of computation times
and f-measure.



\begin{figure}[t]
    \centering
    \begin{subfigure}[b]{0.495\textwidth}
        \centering
        \includegraphics[width=\textwidth]{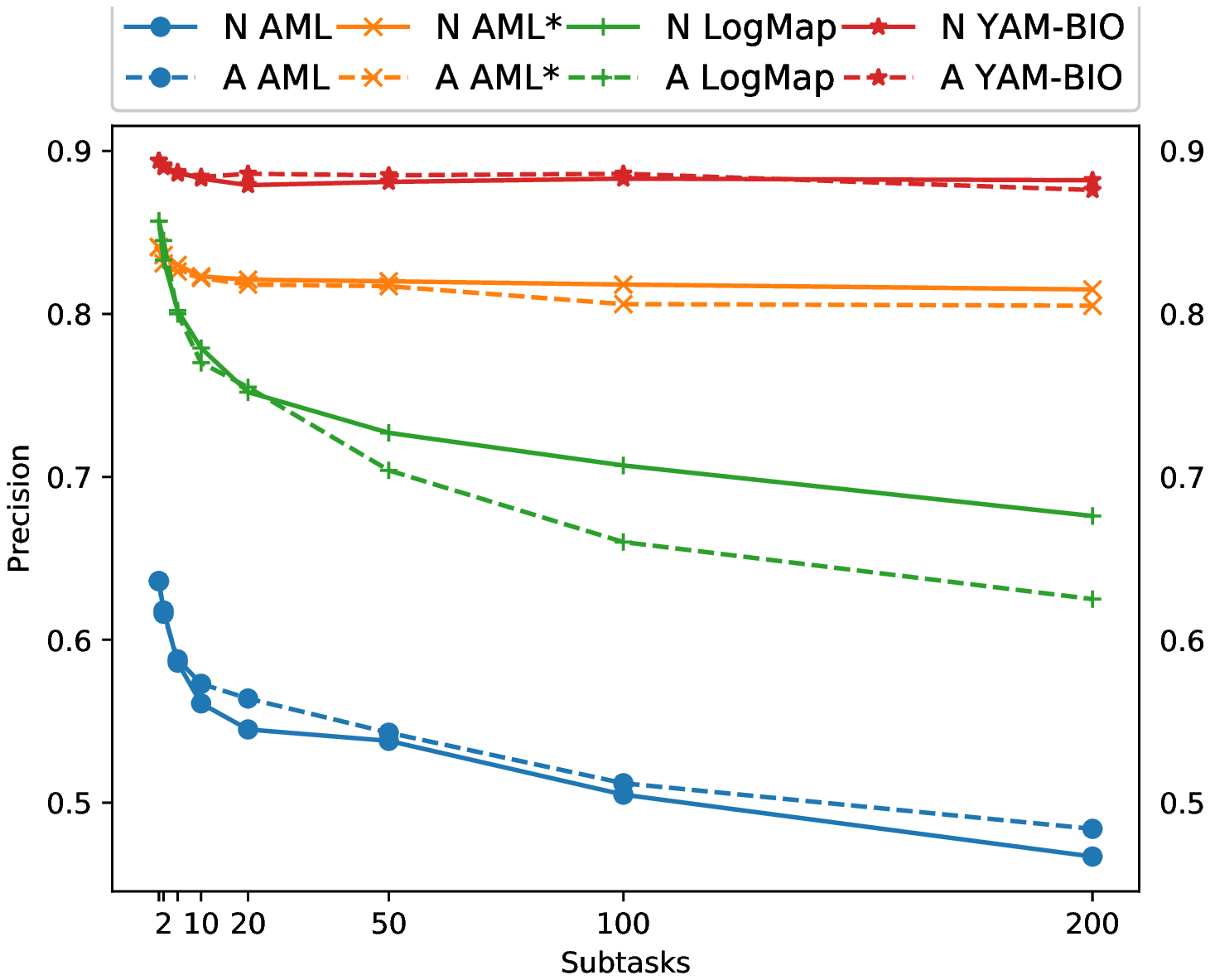}\\[-1ex]
        \caption{Precision.}%
        \label{fig:precision}
    \end{subfigure}
    \begin{subfigure}[b]{0.495\textwidth}
        \centering
        \includegraphics[width=\textwidth]{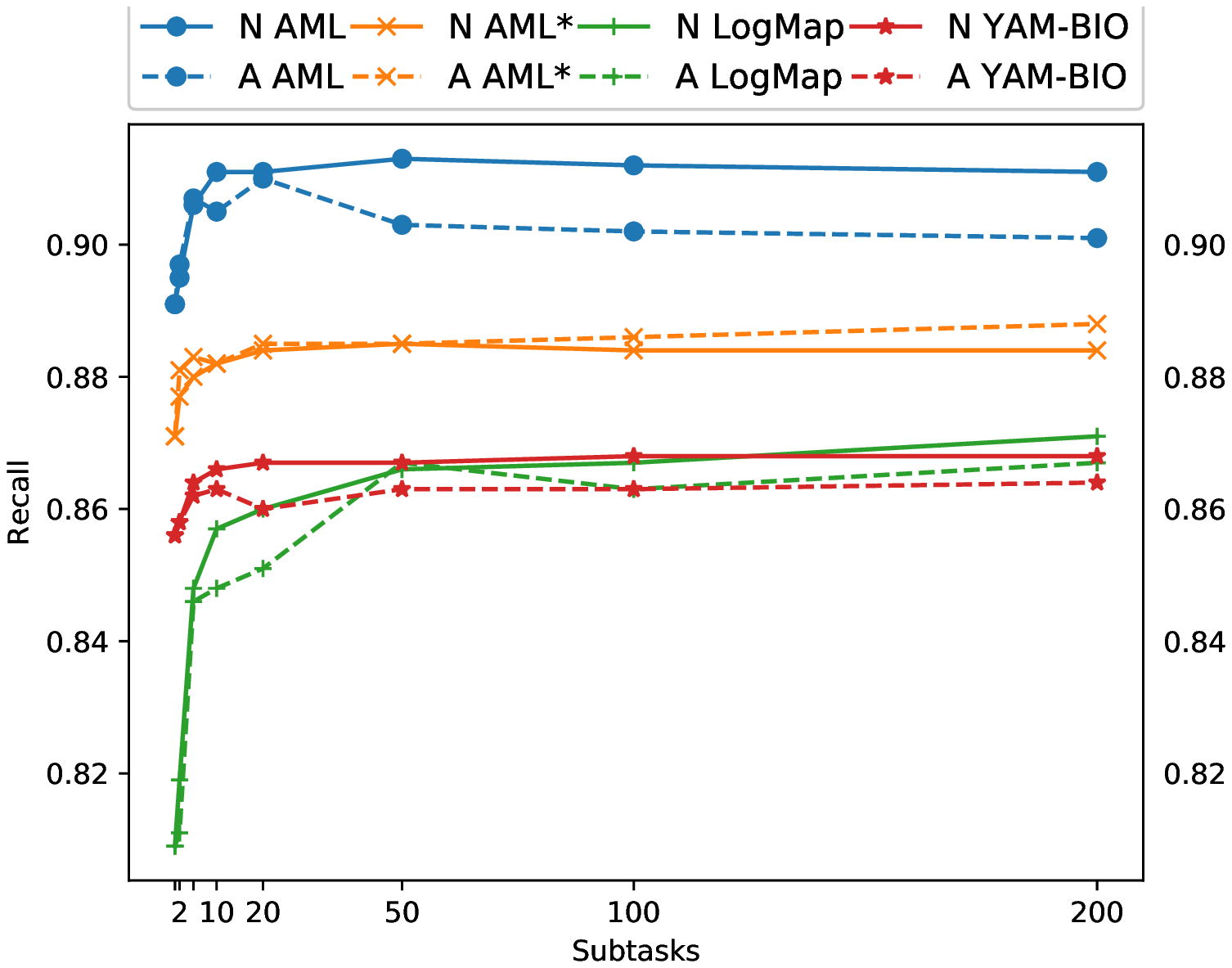}\\[-1ex]
        \caption{Recall.}%
        \label{fig:recall}
    \end{subfigure}
   	\\[1.5ex]
    \begin{subfigure}[b]{0.495\textwidth}
        \centering
        \includegraphics[width=\textwidth]{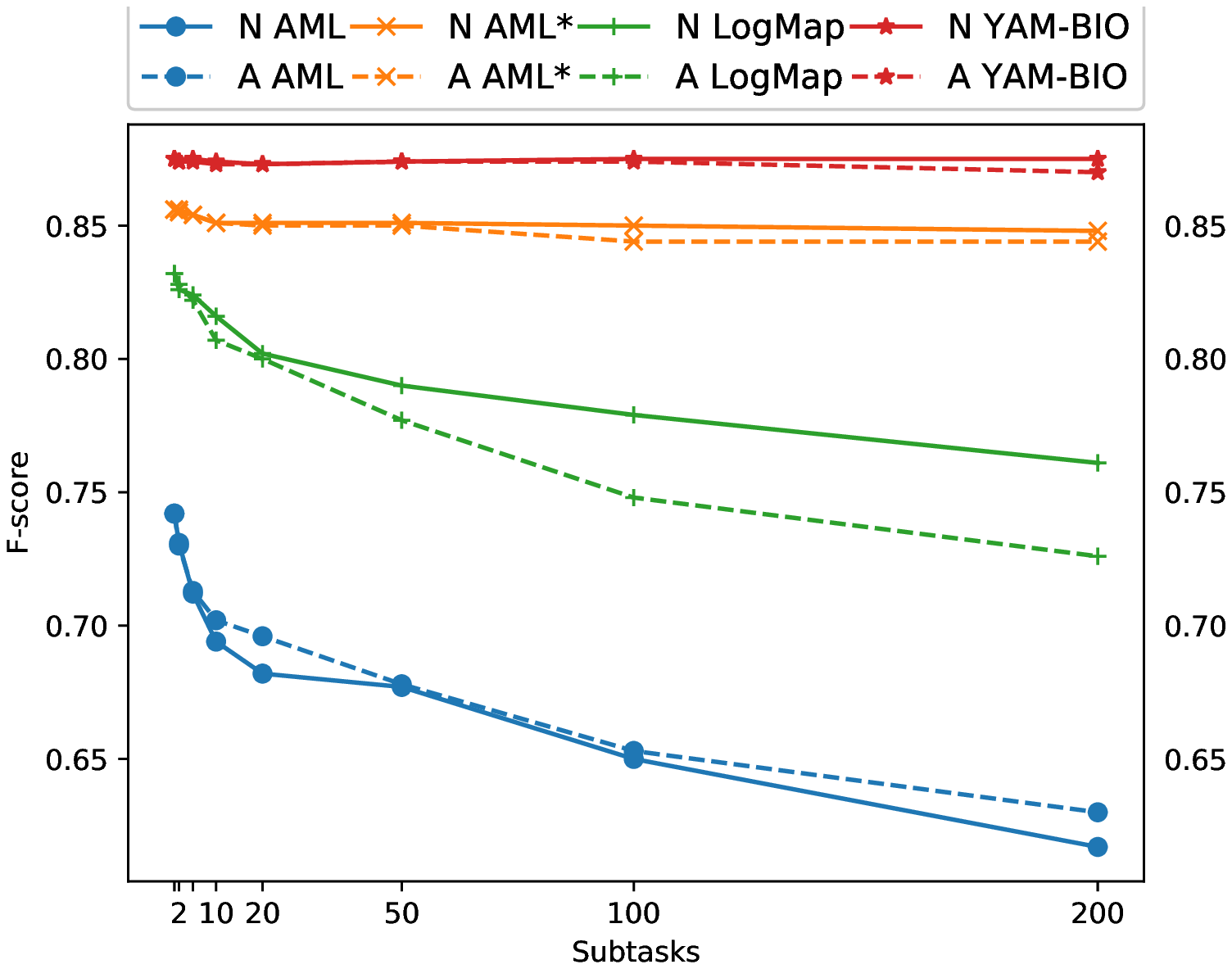}\\[-1ex]
        \caption{F-measure.}
        \label{fig:fscore}
    \end{subfigure}
    \begin{subfigure}[b]{0.495\textwidth}
        \centering
        \includegraphics[width=\textwidth]{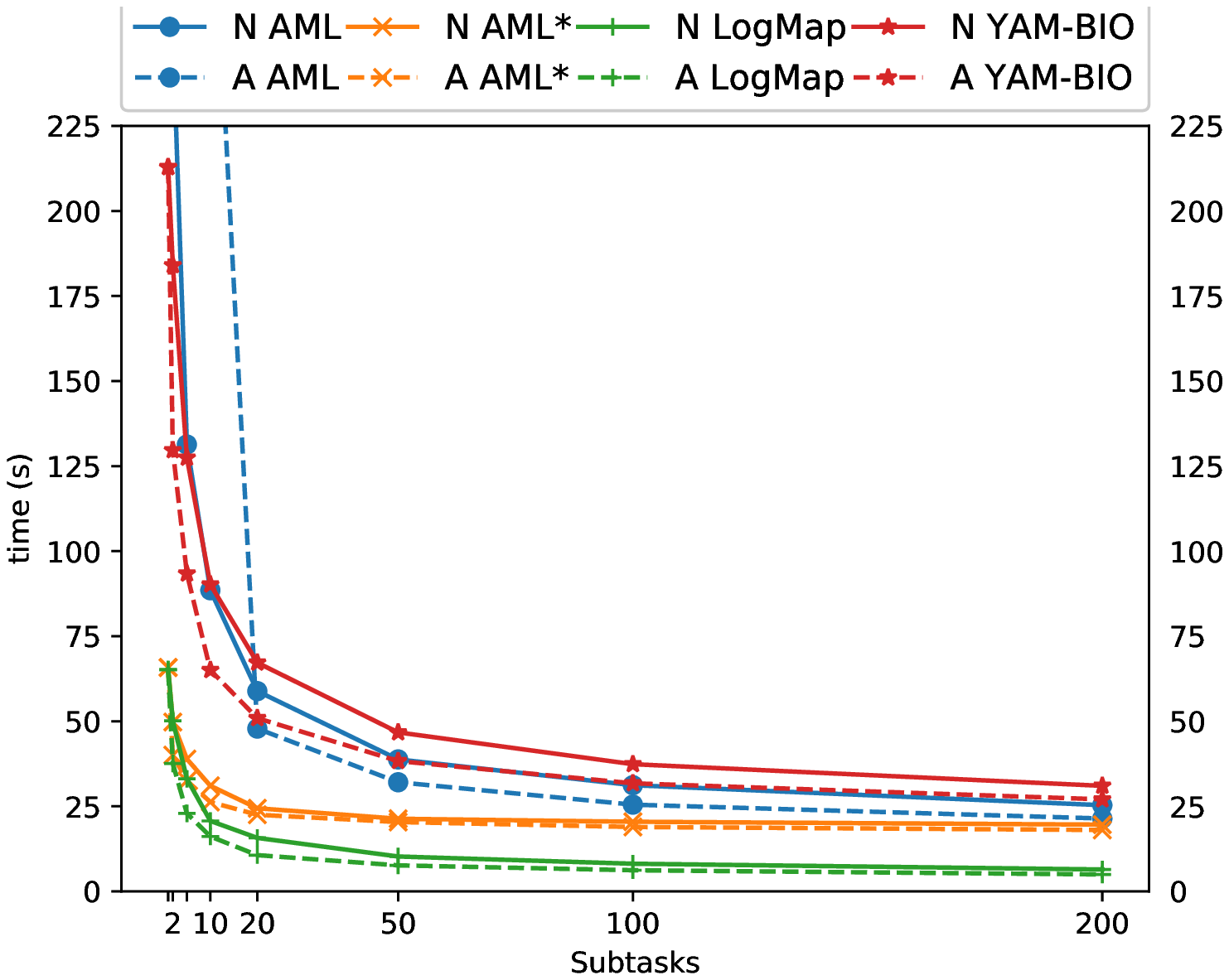}\\[-1ex]
    	\caption{Time (s) per matching task.}
        \label{fig:time}
    \end{subfigure}
    \caption{Performance of top-systems in FMA-NCI task for the divisions
    $\PMT^n$. Original OAEI 2017 results: YAM-BIO \textit{(P: 0.82, R: 0.89, F:
    0.85, t: 279s)}, AML \textit{(P: 0.84, R: 0.87, F: 0.86, t: 77s)}, LogMap
    \textit{(P: 0.86, R: 0.81, F: 0.83, t: 92s)}.}
    \label{fig:performance}
\end{figure}

\vspace{-0.1cm}
\paragraph{Performance of top OAEI systems.} 
We have also evaluated the top systems in the OAEI 2017 
 \emph{largebio} track \cite{DBLP:conf/semweb/AchichiCDEFFFFH17} (LogMap, AML
 and YAM-BIO) to \begin{inparaenum}[\it i)]
\item confirm the dismissal of Hypothesis \ref{prop:performance},
 and
\item evaluate the effect of the divisions of a matching task in the performance of a system.
\end{inparaenum}
Figure \ref{fig:performance} shows the results for the divisions $\PMT^n$
of FMA-NCI with $n \in \{2, 5, 10, 20, 50, 100, 200\}$. 
Solid lines represent
the results for the divisions computed with the naive strategy while the
 neural embedding strategy results are represented with dashed lines. For example, in
 the figure legends, ``N AML'' stands for the results of AML with the divisions
 using the (N)aive strategy while ``A AML'' stands for the results of AML with
 the (A)dvanced (\ie neural embedding) strategy divisions. 
 The results for the naive and neural embedding strategies are very similar,
 with the exception of LogMap, for which results are slightly different for
 $n>50$. YAM-BIO maintains almost constant values for precision, recall
 and f-measure. The f-measure of YAM-BIO is improved with respect to the
 original OAEI results. The results for AML and LogMap are less positive as the
 number of matching subtasks (\ie $n$) increases. Recall increases with $n$ but remains relatively
 constant for $n>50$, however precision is highly impacted by~$n$.
 These results  weaken the validity of Hypothesis
 \ref{prop:performance}.
 The decrease in LogMap's precision may be explained by the fact that LogMap
 limits the cases of many to many correspondences and, when the alignment task
 is divided, that filter is probably not triggered leading to an increase of
 the false positives. Regarding AML performance, we contacted the developers of AML to get
 a better insight about the results. For the type and size of the computed matching tasks AML automatically
 applies a less conservative matching pipeline which leads to an increase of the
 recall, but also to a notable decrease in precision. We also evaluated AML
 forcing a (conservative) pipeline (referred to as AML* in Figure
 \ref{fig:performance}). AML* obtains the expected results, which are very
 similar to the original OAEI results for all the divisions $\PMT^n$.
 The times reported in Figure \ref{fig:time} represent averages per matching
 task. The times for AML were also higher than expected.
 As expected the necessary time to complete a task is reduced with $n$.
 The total required time, however, is increased for all three evaluated
 systems. For example LogMap requires around 100s to complete the two
 matching tasks in $\PMT^2$, while it needs more than 800s to complete the
 $100$ matching 
 tasks in $\PMT^{100}$.  This is explained by the fact that these systems
 implement efficient indexing and matching  techniques and a large portion of the
 execution time is devoted to loading, processing and initialization of the
 matching task.
 Nevertheless, if several tasks are run in parallel, the wall-clock times
 can be reduced significantly. For example, the HOBBIT platform adopted
 for the OAEI 2017.5 and 2018 evaluation campaigns includes 16 hardware cores devoted for the
 system evaluation \cite{hobbit16}. Thus, total 
 wall-clock times could potentially be split~by~16.

\vspace{-0.1cm}
\section{Related work}
\label{sec:related}

\vspace{-0.01cm}
The use of partitioning and modularization techniques have been extensively used
within the Semantic Web to improve the efficiency when solving the task at
hand (\eg ontology visualization \cite{stuckenschmidt:2009, agibetov_2015}, 
ontology reuse \cite{DBLP:conf/esws/Jimenez-RuizGSSL08}, ontology debugging
\cite{DBLP:conf/aswc/SuntisrivarapornQJH08}, ontology classification
\cite{DBLP:conf/semweb/RomeroGH12}).
Partitioning has also been widely used to reduce the complexity of the ontology
alignment task. In the literature there are two major categories of partitioning
techniques, namely: \emph{independent} and \emph{dependent}. Independent
techniques typically use only the structure of the ontologies and are not concerned about
the ontology alignment task when performing the partitioning. Whereas
dependent partitioning methods rely on both the structure of the ontology and
the ontology alignment task at hand. Our approach, although we do not compute (non-overlapping) partitions of the ontologies, can be considered a type of 
dependent technique.

Prominent examples of ontology alignment systems including partitioning
techniques are Falcon-AO \cite{hu:2008}, COMA++ \cite{alger:2011} and TaxoMap
\cite{Hamdi:2010}. 
Falcon-AO and COMA++ perform independent partitioning where the clusters
of the source and target ontologies are independently
extracted. Then pairs of similar clusters (\ie matching subtasks) are
aligned using standard techniques. TaxoMap \cite{Hamdi:2010} implements a
dependent technique where the partitioning is combined with the matching
process. TaxoMap proposes two methods, namely: PAP (partition, anchor,
partition) and APP (anchor, partition, partition). The main difference of these
methods is the order of extraction of (preliminary) anchors to discover pairs of 
partitions to be matched (\ie matching subtasks).

Algergawy et al. \cite{Alger:2015} have recently presented SeeCOnt, which
proposes a seeding-based clustering technique to 
discovers independent
clusters in the input ontologies. 
Their approach has been 
evaluated 
with the Falcon-AO
system by replacing its native PBM (Partition-based Block Matching) module~\cite{pbm2016}.

The above approaches, although they present interesting results,
did not provide any guarantees about the coverage (as in
Definition~\ref{def:cov2}) of the discovered partitions or divisions.~In~\cite{Pereira:2017}
we performed a preliminary study with the PBM method of Falcon-OA, and the PAP
and APP methods of TaxoMap. The results in terms of coverage
with the \emph{largebio} tasks were very low, which directly affected
the results of the evaluated systems. These rather negative results encouraged
us to work on the approach presented in this~paper. 

Our dependent approach, unlike traditional partitioning methods, computes
overlapping self-contained modules (\ie locality modules). 
Locality modules guarantee the extraction of all semantically related entities for a given signature, which enhances the coverage results and enables the inclusion of the relevant information required by an alignment system.
It is worth mentioning that 
the need of self-contained and covering modules, although not thoroughly
studied, was also highlighted in a preliminary work by
Paulheim~\cite{Paulheim:2008}.



\section{Conclusions and future work}
\label{sec:disc}

We have developed a novel framework to split the ontology alignment
task into several matching subtasks based on a lexical index and
locality modules. These independent matching subtasks can be potentially run
in parallel in evaluation platforms like the HOBBIT \cite{hobbit16}. 
%
We have also presented
two clustering strategies of the lexical index.
One of them relies on a simple splitting method, while the other relies on a
fast (log-linear) neural embedding model.
We have
performed a comprehensive evaluation of both strategies which suggests that
the  obtained divisions are suitable in practice in terms of both coverage and size. 
The naive strategy leads to well-balanced set of tasks, while the overall
reduction of the search space with the neural embedding strategy was very
positive.
%
The division of the matching task also allowed us to obtain results for
five systems which failed to complete these OAEI matching tasks in the past.

The results in terms of f-measure were not as good as expected for some of the
systems. The f-measure also tended to decrease as the number of matching subtasks increased. These results, although not supporting our original intuitions, do not undermine the value of the proposed framework as we cannot control the internal behaviour of the ontology alignment system.
%
Computed matching subtasks for a given division $\PMT^n$ may have a high
overlapping, especially when relying on the naive strategy. That is, the same
mapping can be proposed from different matching subtasks. This can
enhance the discovery of true positives, but may also bring in a number of
false positives, as for the case of LogMap in the reported
evaluation.
The adoption of the presented framework within the
pipeline of an ontology alignment system may also lead to improved results,
as for the case of YAM-BIO and AML with a conservative pipeline. It is worth
mentioning that the 
OAEI system SANOM (v.2018) is already
integrating the strategies presented in this paper within its matching workflow.


Both the naive and the neural embedding strategies require the size of the
number of matching subtasks or clusters as input. The (required) matching subtasks may be known before
hand if, for example, the matching tasks are to be run in parallel in a number
of available CPUs. For the cases where the resources are limited or where a
matching system is known to cope with small ontologies, we plan to design an
algorithm to estimate the number of clusters so that
the size of the matching subtasks in the computed divisions is appropriate to the system and resource constraints.
As immediate future we also plan to study different notions of \emph{context}
of an alignment (\eg the tailored modules proposed in
\cite{DBLP:journals/jair/RomeroKGH16}).
Locality-based modules, although they have led to very good results,
can still be large in some cases.


\subsubsection*{Acknowledgements.}
EJR was funded by the Centre for Scalable Data Access (SIRIUS), the RCN project BigMed, and The Alan Turing project AIDA. We would also like to thank the anonymous reviewers that helped us to improve this contribution.


\bibliographystyle{splncs}
\bibliography{references}  

\end{document}